%% file: main.tex
\documentclass{article}

\PassOptionsToPackage{sort&compress,numbers}{natbib}

\usepackage[preprint]{neurips_2022}

\usepackage[utf8]{inputenc} 
\usepackage[T1]{fontenc}    
\usepackage{hyperref}       
\usepackage{url}            
\usepackage{booktabs}       
\usepackage{amsfonts}       
\usepackage{nicefrac}       
\usepackage{microtype}      
\usepackage{xcolor}         

\usepackage{siunitx}
\DeclareSIUnit\pixel{px}

\usepackage{subcaption}     
\usepackage{amsmath}
\usepackage{placeins}
\usepackage{graphicx}

\usepackage{bm}

\title{Recurrent networks improve neural response prediction and provide insights into underlying cortical circuits}

\author{
  Yimeng Zhang \\
   \And
    Harold Rockwell \\
    \And
    Sicheng Dai \\
    \And
    Ge Huang \\
   \And
    Stephen Tsou \\
    \And
    Yuanyuan Wei
   \And
   Tai Sing Lee \\
    Computer Science Department and Neuroscience Institute\\
    Carnegie Mellon University \\
    Pittsburgh, PA 15213 \\
    \texttt{tai@cnbc.cmu.edu}
}

\begin{document}

\maketitle

\begin{abstract}
  Feedforward CNN models have proven themselves in recent years as state-of-the-art models for predicting single-neuron responses to natural images in early visual cortical neurons. In this paper, we extend these models with recurrent convolutional layers, reflecting the well-known massive recurrence in the cortex, and show robust increases in predictive performance over feedforward models across thousands of hyperparameter combinations in three datasets of macaque V1 and V2 single-neuron responses. We propose the recurrent circuit can be conceptualized as a form of ensemble computing, with each iteration generating more effective feedforward paths of various path lengths to allow a combination of solutions in the final approximation. The statistics of the paths in the ensemble provide insights to the differential performance increases among our recurrent models.  We also assess whether the recurrent circuits learned for neural response prediction can be related to cortical circuits. We find that the hidden units in the recurrent circuits of the appropriate models, when trained on long-duration wide-field image presentations, exhibit similar temporal response dynamics and classical contextual modulations as observed in V1 neurons. This work provides insights to the computational rationale of recurrent circuits and suggests that neural response prediction could be useful for characterizing the recurrent neural circuits in the visual cortex.
\end{abstract}

\graphicspath{{figures/}}

\input{shared.tex}

\input{intro.tex}

\input{methods.tex}

\input{results.tex}

\input{discussion_new.tex}
\bibliographystyle{unsrtnat}
\bibliography{all_combined}

\clearpage

\section{Appendix}
\appendix

\input{appendix.tex}

\end{document}

%% file: shared.tex
\newcommand{\vect}[1]{\vec{\boldsymbol{#1}}}
\newcommand{\mat}[1]{\boldsymbol{#1}}
\newcommand{\todo}[1]{\textbf{\color{red} #1}}

\newcommand{\ccabs}{\operatorname{CC}_{\text{abs}}}
\newcommand{\ccmax}{\operatorname{CC}_{\text{max}}}
\newcommand{\ccnorm}{\operatorname{CC}_{\text{norm}}}

\newcommand{\ccraw}{\operatorname{CC}_{\text{raw}}}

\newcommand{\nnconv}{\operatorname{Conv}}
\newcommand{\nncpb}{\operatorname{CPB}}
\newcommand{\nnrcpb}{\operatorname{RCPB}}
\newcommand{\nnavgpool}{\operatorname{Pool}}
\newcommand{\nnbn}{\operatorname{BatchNorm}}
\newcommand{\nnact}{\operatorname{Act}}
\newcommand{\nnfc}{\operatorname{FC}}

%% file: intro.tex
\section{Introduction}
\label{nips2021:intro}

Feed-forward deep neural networks have been shown to be effective models for modeling and predicting neural responses of early visual areas in the brain \citep{10.1167/19.4.29,DBLP:conf/nips/KlindtEEB17,Yamins:2016hg,Zhang2019,10.1371/journal.pcbi.1006897,doi:10.1146/annurev-vision-082114-035447}. However, abundant recurrent connections exist within and between visual areas and are crucial to generating neural responses  \citep{10.1093/cercor/1.1.1-a,10.1002/cne.23458}.  While biologically-inspired recurrent circuits incorporated in convolution neural networks have been shown to be more efficient than purely feedforward models, accomplishing comparable performance in image classification tasks with less layers and less parameters \citep{DBLP:conf/nips/HanWZFCL18,nayebi2018task,DBLP:conf/nips/KubiliusSHMRIKB19},  their underlying computational rationales remains to be understood.  In this paper, we investigated the use of recurrent circuits in neural system identification with the following motivating questions:  (1)  Since neural data in the visual cortex are generated by recurrent circuits, would the use of a similar architecture provide a better model for predicting neural response?  (2) What are the computational rationales and network features responsible for the performance advantages of recurrent circuits? (3)  Finally, could neural response prediction using a network model with recurrent circuits provide a useful approach for characterizing and understanding the underlying recurrent neural circuits in the  cortex?   

Our task is to perform neural system identification by training neural network models with recurrent circuits to predict the mean firing rates of recorded macaque early visual cortical neurons (V1 and V2) in response to natural input images, when the animals were performing a fixation task. We separate the model into two stages, one stage including feedforward convolution layers and recurrent circuits to generate the inputs for the target neuron to be modeled, and a second stage, called readout, that models processing by the target neurons and temporal average of their responses to generating the target signals to be predicted. Three sets of data are tested, with distinct stimulus sizes and presentation duration (see Section A and Figure 1 of Appendix for details), to characterize the conditions under which recurrent circuits could exhibit performance gain over feedforward circuits. 
Neural response prediction models are sensitive to hyperparameters. To establish the computational advantages of the recurrent circuits over feedforward network in general, rather than specific to some optimized hyperparameters and architectural choices, we compare each network with recurrent circuits against its hyper-parameter and parameter matched feedforward model, across a large number of hyperparameters for model size (width and depth), activation functions (ReLU or softplus), model initialization seeds, and the amount of training data. 
In addition, we investigate four readout paradigms designed to model how the targeted neurons process the temporal responses of the hidden units from the recurrent circuits as well as the computation involved in creating the targeted signals for the network to fit. 


Our contributions are three-fold. First, our experimental results conclusively show that recurrent networks perform better than hyperparameter-matched feedforward networks for predicting the responses of neurons in the early visual cortex. This performance gain is general, and not specific to a variety of hyperparameter choices, but more significant for more limited data size. This performance gain could arise from at least two possible sources:  recurrent network (1) being a better or more efficient function approximator or (2) being a better match of the underlying circuitry in the early visual cortex, or both. Our second contribution  addresses the first possibility by proposing that recurrent circuits can be conceptualized as a multi-path ensemble, reaping the benefits of ensemble computing \cite{opitz1999,Polikar2006,rokach2010}. We demonstrated empirically multi-path ensemble model approximates well the performance of recurrent circuits. This supports the conceptualization but also allows us to dissect the recurrent circuits from the multi-path ensemble perspective. Our third contribution addresses the second possibility by showing that the hidden units of the recurrent circuits exhibit response temporal dynamics and contextual modulation observed in early cortical neurons even when the training stimuli and targeted signals were not designed to create these dynamics and contextual modulation effects. 
Our findings provide fresh insights to the rationales underlying the computational advantages of recurrent circuits in neural networks and possibly in the brain, and suggest that the response prediction paradigm potentially could be useful for characterizing cortical recurrent circuits in the brain.

\section{Related work}

Recent work has demonstrated that adding biologically-inspired recurrent circuits to feedforward CNNs can lead to performance comparable to that of the state-of-the-art in image classification, but with significant less model parameters \citep{nayebi2018task,DBLP:conf/nips/KubiliusSHMRIKB19,10.1371/journal.pcbi.1008215,DBLP:conf/nips/HanWZFCL18}.  
This suggests that recurrent circuits may provide a better or at least more efficient functional approximation. Here, we consider this approach for tackling the problem of neural response prediction. Rather than seeking state-of-the-art models or performance, we seek to assess whether recurrent networks are better than hyperparameter-matched feedforward models across many hyperparameters -- where each recurrent layer is equated to two feedforward layers, to study the performance gain of recurrent computation in a more pure form.

 
In earlier work that incorporated recurrent circuits into CNNs for image classification, it was also found that the recurrent hidden units can exhibit temporal response dynamics reminiscent of that IT neurons \cite{nayebi2018task} and the learned weight matrices of the recurrent kernels exhibit patterns that might give rise to known physiological responses \cite{10.3389/fpsyg.2017.01551}. These works however did not  explicitly perform "neurophysiological experiments" on the hidden units to see if the neurons actually exhibit classical contextual modulation that  presumably emerge from computation in recurrent circuits. Inspired by the works of \cite{McIntosh:2017uc} on retina, we performed some neurophysiological tests on the hidden units in the recurrent circuits trained to fit V1 and V2 responses.  Our work differs from \cite{McIntosh:2017uc} in that we fit network to predict mean response to static images, rather than temporal response to dynamic video. While it is possible to use a feedforward network to predict temporal responses to  static images, it inherently puts the feedforward network to a  competitive disadvantage relative to the matched recurrent network which has intrinsic dynamics. It is more compelling to let recurrent network compete in a problem in which feedforward network has been shown to be able to achieve the state-of-the-art performance,  than in a problem where a feedforward network is not known to do well. Interestingly, even when we trained the network to fit mean responses, we found the hidden units exhibit response dynamics that are similar to what we observed in V1 and V2 neurons to static images.

 
Although recurrent networks have been shown to produce performance improvements in a number of goal-driven tasks, the theoretical reasons underlying their success are not well understood.  Deep equilibrium theory \citet{DBLP:conf/nips/BaiKK19}, showing weight-tied deep recurrent networks converge to a fixed point of the recurrent dynamics,  provides an important perspective, but it has not helped us understand performance of the recurrent circuits in our setting. 
Here, we propose a different perspective, in the form of multi-path ensemble computing, to reason about recurrent computation. Treating recurrent computation as a multi-path ensemble is novel even though multi-path ensembles have been evoked to explain deep networks with residual connections \citep{DBLP:conf/cvpr/HuangLMW17,NIPS2016_37bc2f75}, \citep{DBLP:journals/corr/SrivastavaGS15}, \citep{DBLP:conf/cvpr/HeZRS16,DBLP:conf/eccv/HeZRS16}, \citep{DBLP:conf/iclr/LarssonMS17}, and DenseNet \citep{DBLP:conf/cvpr/HuangLMW17}.  \citet{DBLP:journals/corr/LiaoP16} did suggest the equivalence between a specific type of recurrent networks with weight-tied ResNets, and \citet{DBLP:conf/nips/ChenRBD18} formulated deep weight-tied ResNets to approximate the dynamics of continuous recurrent models specified in ordinary differential equations.
Our work explicitly connects recurrent models to multi-path ensemble models and explores the use of this reformulation to characterize recurrent networks in terms of path length statistics to understand the differential performance of the different models.

%% file: methods.tex
\section{Methods}
\label{nips2021:method}

\subsection{Comparison between the recurrent circuit models and feedforward models in neural response prediction} 
Our first objective is to evaluate whether and what recurrent network models perform better in predicting the mean firing rate of individual neurons than feedforward networks of matched model parameter sizes and hyperparameters.

{\bf Datasets:}
In this neural system identification problem, We train each model on three datasets independently. All three sets contains V1 and V2 neurons' responses to natural images recorded using multi-electrode arrays from macaque monkeys performing a fixation task. We call them the {\bf RF-S}, {\bf WF-S}, and {\bf WF-L} datasets.  RF-S's stimuli were presented in an  aperture close to the size of the classical receptive field (RF) and in short (S) duration (60 ms) each in rapid succession. The {\bf RF-S} data are public-domain data under a Creative Commons license, from \cite{10.1371/journal.pcbi.1006897} contains 115 V1 neurons' responses to 7250 images. The {\bf WF-S} and {\bf WF-L} data were collected by us, containing 79 and 34 V1/V2 neurons responding to 8000 and 2250 images respectively in response to stimuli presented in $8^{o}$ and $10^{o}$ visual angle aperture (wide-field ({\bf WF})) respectively. The presentation duration of each stimulus is 47 ms for the {\bf WF-S} dataset, using the same paradigm as in \cite{10.1371/journal.pcbi.1006897}), and 500 ms for the {\bf WF-L} dataset. Each stimulus was repeated about 10 times in both of these sets. 
These datasets are characterized by different extents of spatial and temporal contextual modulation.  {\bf WF-L} involves a large spatial context and long temporal processing for each image, thus significant contextual modulation is expected. {\bf WF-S} involves extensive spatial context but does not allow much temporal processing. {\bf RF-S}, with minimal spatial context and temporal integration, is expected to involve the least amount of contextual modulation. We expect the recurrent circuits learned will strongly depend on the degree of contextual modulation. Note that the {\bf WF-L} and {\bf WF-S} datasets include both V1 and V2 neurons. 



\begin{figure}[htb]
   \centering
    \includegraphics[width=\textwidth]{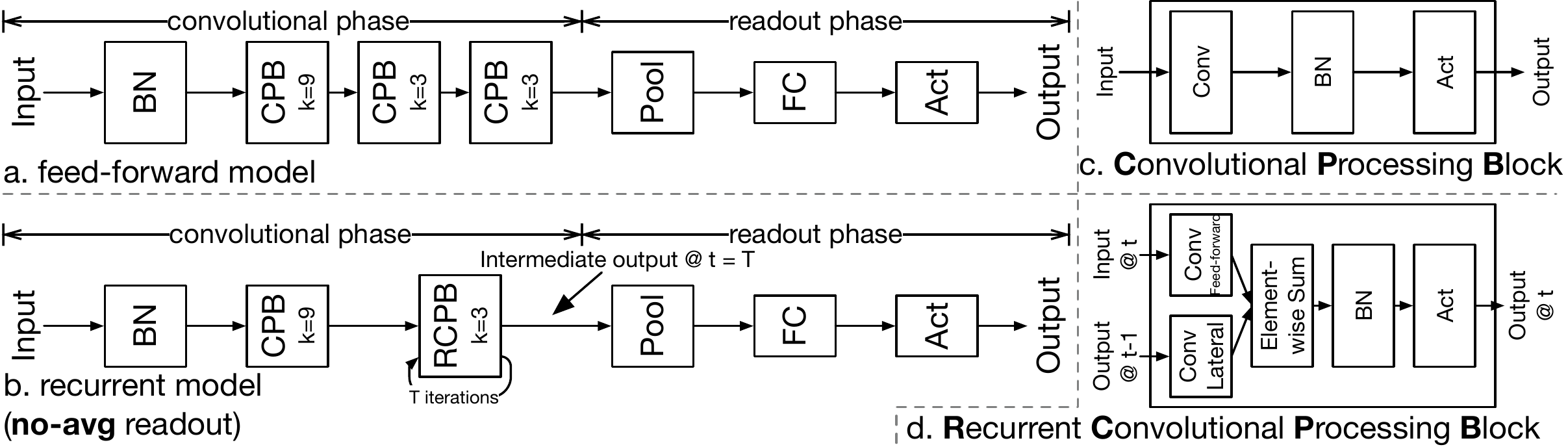}
    \caption{Models explored in this study. \textbf{(a)} shows an example baseline feedforward  model. \textbf{(b)} shows an example recurrent model under the \texttt{no-avg} readout mode.  \textbf{(c)}, \textbf{(d)} show a convolutional processing block (\texttt{CPB}) and a recurrent convolutional processing block (\texttt{RCPB}), respectively. The kernel size of a (R)CPB is denoted by \texttt{k} in \textbf{(a)},\textbf{(b)}. In this figure, we have in total three convolutional blocks for the feedforward model \textbf{(a)} but only two blocks for the recurrent one \textbf{(b)} because a RCPB equals two CPBs in terms of model size; throughout this study we kept the first layer of a recurrent model to be feedforward per pilot experiments.} 
    \label{fig:methods:datadriven_k_models}
\end{figure}

{\bf Feedforward Baselines:}
Our baseline feedforward models are based on those in \citet{10.1371/journal.pcbi.1006897,DBLP:conf/nips/KlindtEEB17} (Fig.~\ref{fig:methods:datadriven_k_models}a) which are state of the art in predicting neural responses of early visual areas. These models consist of a "shared core" of convolutional layers used for the whole dataset, followed by a factorized readout for each individual neuron. We trained feedforward models of many different model sizes and hyperparameters; each baseline model is a feedforward model (Fig.~\ref{fig:methods:datadriven_k_models}a) that contains a batch normalization  \citep{DBLP:conf/icml/IoffeS15} layer, a set of convolutional processing blocks (CPBs), as well as pooling, fully connected, and activation layers. The hyperparameters varied include amount of training data (25\%, 50\%, and 100\%), activation function (ReLU or Softplus), model size (number of layers/channels), and others, resulting in at least 288 feedforward models for each dataset. 


{\bf Recurrent Models:} For each FF model, we will compare it with recurrent networks (with the same hyperparameters, and parameters) with the simplest recurrent interaction, in which a weighted sum (by a lateral interaction kernel) of all the neurons in a neuron's 3 x 3 neighborhood at each iteration is added to its response in the previous iteration. Since the CPB's kernel is 3x3xk where k is the number of feature channels, the lateral interaction kernel is thus also 3x3xk. From the perspective of model size, each Recurrent Convolutional Processing Block (RCPB) in the recurrent model  is equivalent in the number of parameters to two convolution processing blocks (CPB) in the FF baseline model as shown in (Fig.~\ref{fig:methods:datadriven_k_models}b,d). A recurrent convolutional processing block (RCPB) extends regular CPB by having two inputs: a \emph{bottom-up} input from the previous stage of the model in the current iteration and a \emph{lateral} input from the output of the RCPB in the previous iteration (Fig.~\ref{fig:methods:datadriven_k_models}d).   
The batch normalization parameters are learned separately for each inference iteration, following the design of \citet{10.1371/journal.pcbi.1008215}. We experimented with networks with one RCPB and with two RCPBs.  Two-RCPB models are parameter-matched to feedforward models with four CPBs. During model inference, information flows over each CPB only once, whereas it flows over each RCPB for a certain number of iterations. The number of iterations allowed is a model hyperparameter; when the number of model iterations is 1, a recurrent model is effectively a feedforward one. For each  FF model and for each type of recurrent model architecture type,  we trained six recurrent models,  one for each number of iterations, with the number of iterations range from 2 to 7. Note that the model with 1 iteration is effectively a feedforward model but with less parameters than the feedforward model being compared, and thus will not be included in the comparison.

{\bf Variations in Architecture and Readout Modes:} 
For both feedforward and recurrent models, the model inference can be divided into two phases:  convolutional and readout phases (Fig.~\ref{fig:methods:datadriven_k_models}). 
The convolutional phase contains a BN layer followed by (recurrent) convolutional processing blocks, each of which contains convolution, BN, and activation layers; the readout phase takes the output of the final convolutional block through average pooling, factorized fully-connected \citep{DBLP:conf/nips/KlindtEEB17}, and activation layers to get the final model output.  The readout phase represents the operation performed by the neuron to be modeled on the bottom-up input signals provided by the convolution phase, as well as the operation performed by the experimenter, averaging the temporal response in our case, to produce the teaching signals to be fitted by the model. 

\begin{figure}[htb]
    \centering
    \includegraphics[width=\textwidth]{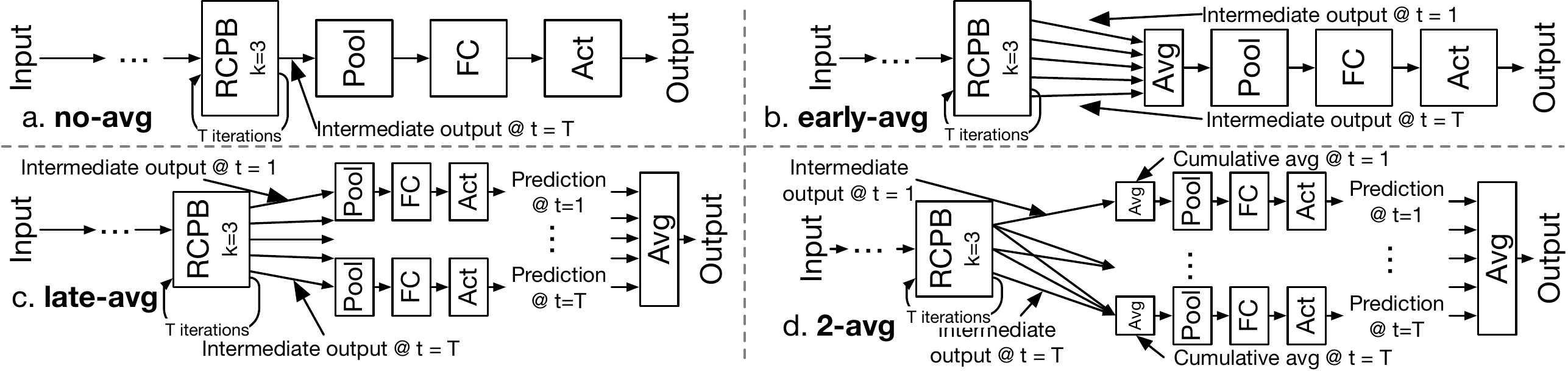}
    \caption{Readout modes \texttt{no-avg} (\textbf{a}),  \texttt{early-avg} (\textbf{b}), \texttt{late-avg} (\textbf{c}), and \texttt{2-avg} (\textbf{d}) explored in this study. The simplest one (\texttt{no-avg}) is also shown earlier in Fig.~\ref{fig:methods:datadriven_k_models}b. The style of the figure follows that of Fig.~\ref{fig:methods:datadriven_k_models} and the convolutional phases which are the same across readout modes are omitted.} 
    \label{fig:methods:readout}
\end{figure}

There are four types of readout paradigms in the readout phase for each of the recurrent models  (Fig~\ref{fig:methods:readout}), and the learnable parameters are the same for the four readout modes, as well as that of the feedforward counterpart.
For a model with $T$ iteration, its RCPB running for $T$ iterations will generate $T$ intermediate outputs. The target neurons integrate and process these outputs differently in the four readout modes.
\texttt{no-avg}, simply uses the output after the $T$ iteration, as typical for convolutional RNNS~\citep{DBLP:conf/nips/KubiliusSHMRIKB19}.  \texttt{early-avg}  averages the  RCPB's outputs across time and feeds the mean to the readout phase after the $T$ iteration, as was done in some bio-inspired RNNs for object recognition~\citep{10.1371/journal.pcbi.1008215}. These two readout modes do not generate a temporal response at the ACT stage of the readout phase. 
In both \texttt{late-avg} and \texttt{2-avg} modes, the target neurons process the output of the RCPB units in "real-time" at every iteration and produce an output at the ACT stage at every iteration, hence generating a temporal response for the targeted neuron. The temporal responses of the targeted neurons are then averaged to produce the mean response to match the teaching signals. Thus the Ave operation at the end models the temporal averaging that had been performed on the neural data. The main difference between them is that the target neurons process the output of the RCPB independently at each iteration in the \texttt{late-avg} case, but process the cumulative average of the output of the RCPB up to that iterations in the  \texttt{2-avg} case. Thus, target neurons in \texttt{2-avg} remembers the history of the RCPB's responses for each stimulus up to that iteration when generating their responses at that time point. 
The four readout modes represent our hypotheses or assumptions on the computations underling the generation of the experimental neural data.  Neural response prediction performance can serve as a metric to evaluate these assumptions. 
 
In summary, for each feedforward model, there are 24 hyperparameter-match recurrent counterparts ( six numbers of iterations (2 to 7), and four readout modes). Thus, for the 288 feedforward models tested for each dataset, there would be 6912 recurrent models to be compared against for each dataset.

{\bf Model training and evaluation:}
\label{nips2021:method:implementation-details:model_training}
For each data set, roughly 64\%, 16\%,and 20\% of images were allocated for training, validation, and testing, respectively. The validation set was used to control early stopping of the models, since we did not select hyperparameters. 
We constructed and trained all models using PyTorch \citep{2019arXiv191201703P} on an in-house computing cluster with 4-GPU (NVIDIA GeForce GTX 1080 Ti or similar) nodes. and followed \citet{DBLP:conf/nips/KlindtEEB17} on data preprocessing and model optimization details. 
Given a trained model, we use average $\ccnorm^2$ \citep{10.3389/fncom.2016.00010,Zhang2019} over all neurons and all testing images (\SI{20}{\percent} of all samples) to quantify the model's performance on each data set. Conceptually, $\ccnorm^2$ measures the fraction of variance explained by the model, with trial-to-trial variance in neural responses discounted. 
To get the $\ccnorm^2$ for a neuron, we first compute the raw Pearson correlation $\ccraw$ between the ground truth time- and trial-averaged neural responses and the model responses, then we divide $\ccraw$ by $\ccmax$, which estimates the maximal Pearson correlation coefficient an ideal model can achieve given the noise in the neural data \citep{10.3389/fncom.2016.00010,Hsu:2004ku}, followed by squaring.


\subsection{Multi-path reformulation of recurrent circuit models }
After demonstrating networks with recurrent circuits in the appropriate readout modes consistently outperform their hyperparameter-matched feedforward counterparts, we develop an approach to characterize and understand the computational rationales and network properties that contribute to the performance gain provided by the recurrent circuits in this task.

We propose that a recurrent circuit can be conceptualized as a multi-path ensemble, with each iteration generating additional and longer paths.  Ensemble computing is known to have the so-called diversity bonus, allowing a network to combine many weak classifiers or approximators to build a stronger classifier or approximators \citep{opitz1999,Polikar2006,rokach2010}.  We proceed to reformulate the recurrent model's recurrent processing blocks (RCPBs) into an approximately equivalent multi-path ensemble with multiple feedforward paths of different lengths. The weights of the kernels across the paths are shared in order to keep the model parameters constant.
Fig.~\ref{fig:results:depth_analysis:schema} shows a one-RCPB network’s information flow path at the first three iterations. With each iteration, the network adds a longer path. Fig.~\ref{fig:results:depth_analysis:schema}b approximates information flow as summations of simpler individual feedforward paths with shared components. 
 Decomposing the operation of the recurrent circuit into  into a multi-path ensemble allow us to dissect the recurrent circuits by eliminating paths as well as compare the path statistics of the different models or readout modes to understand their performance difference. We found two path statistics particularly illuminating. The first is the weighted average lengths of the paths of the multi-path ensemble, and the second is the diversity of the distributions of the weighted path lengths.  Each component box in Fig.~\ref{fig:results:depth_analysis:schema} can be associated with a certain weight, which is essentially a gain factor, representing the ratio of the magnitude of each box’ output over that of its input. 
Parameters learned along different paths have different magnitudes which should be related to  the relative contribution of the different paths to the final output. 
The product of the weights of all the box components along a particular path gives the total weight of that path.  See Appendix~\ref{appendix:model_approximation_multi_path} for details and an example of the weight computation. With these weights, we can compute the effective path length and path  distribution diversity of the multi-path ensemble for the different model architectures. 

\subsection{Evaluation of response dynamics and contextual modulation in the hidden units}
\label{methods:neurophys}
Our third objective is to determine the extent to which the recurrent circuits learned by our model produce response properties known to belong to recurrent circuits in visual cortex. We evaluated three features of the recurrent circuits learned by our recurrent models: their average temporal dynamics, the extent of end-stopping they show, and the extent of surround suppression they show. 
Neurons in early visual cortex typically show a rapid, high responses to the onset of a static stimulus, which then decays and plateaus at a lower level~\citep{Zipser7376}. V1 neurons additionally are known to exhibit end-stopping (suppression by oriented bars longer than the preferred length)~\citep{doi:10.1152/jn.1965.28.2.229} and surround suppression (suppression by sine-wave gratings larger than the preferred size)~\citep{doi:10.1146/annurev.ne.08.030185.002203}. 
We present oriented bars and gratings of varying sizes to our trained recurrent models and extracted the responses of centrally-positioned hidden units in the RCPBs, whose place in the recurrent circuit corresponds to superficial neurons in V1 or V2. We then examine the temporal dynamics of these responses, and their tuning to the length of bars and size of gratings (at optimal spatial frequencies and orientations), averaged over 150-200 units across several hyperparameter values. See Appendix~\ref{appendix:neurophys} for more details.

%% file: results.tex
\section{Results}
\label{results}
\subsection{Recurrent models outperformed similarly-sized feed-forward models}
\label{results:datadriven_models}

\begin{figure}[htb]
    \centering
    \includegraphics[width=\textwidth]{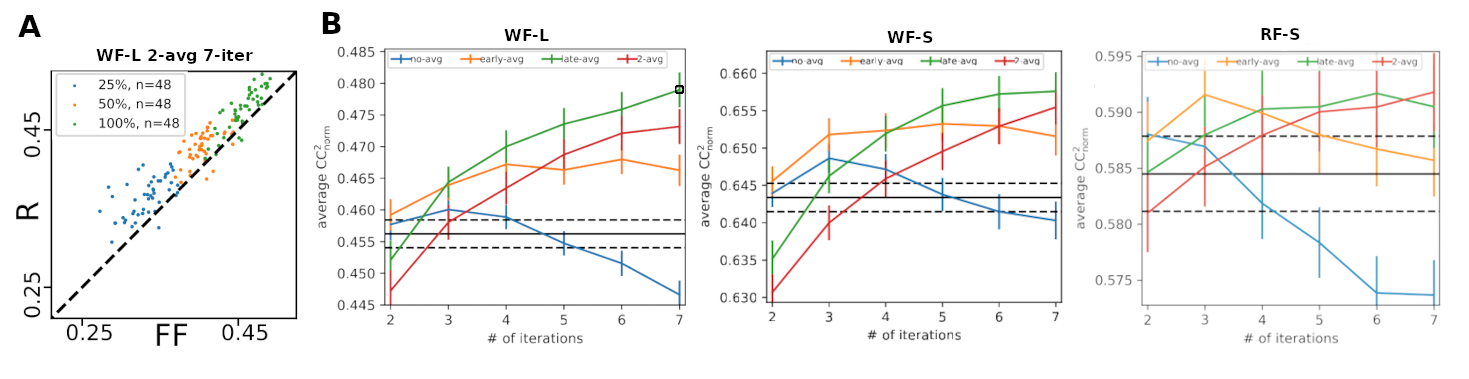} 
    \caption{Recurrent models vs hyperparameter-matched feedforward models on each dataset. Panel {\bf A} illustrates, for 7-iteration \texttt{2-avg} models on the {\bf WF-L} data (circled in panel {\bf B}), the consistent improvement of recurrent models over feedforward models across hyperparameters; each point represents the average performance of two seeds of a model with a particular hyperparameter combination. Panel {\bf B} summarizes this data, with each point in each plot corresponding to the average y-axis value among all the green points in Panel {\bf A}. For all datasets, the \texttt{late-avg} and \texttt{2-avg} readout modes with a high number of iterations consistently outperform parameter-matched feedforward models, and show a clear trend of performance gain for models with larger number of iterations (while the \texttt{no-avg} readout mode decreases).} 
    \label{fig:results:tang:scatter_r_vs_ff:3rd_nips}
\end{figure}

For models trained for each of the three datasets, we compare the feedforward models with their 24 types of parameter-matched recurrent counterparts (4 readout modes, 6 iteration numbers). 
Figs.~\ref{fig:results:tang:scatter_r_vs_ff:3rd_nips}B shows three plots. Each plot summarizes the performance of the 24 model types for each dataset, trained with 100\% training data. The averaged performance of all the tested feedforward models is indicated by the horizontal solid line with the dotted line indicating 1 standard error. The readout modes are indicated by different colors.  The mean performance of 48 recurrent models of this type  of models with 7-iteration and 2-ave read-out mode is circled in the WF-L plot, with the vertical bar indicating standard error. The performances of the 48 recurrent models correspond to the 48 green dots in 
Fig.~\ref{fig:results:tang:scatter_r_vs_ff:3rd_nips}A. Each green dot compares the performance of one recurrent model of this type, trained on 100\% of the training set, against its parameter and hyperparameter-matched feedforward model, i.e. the recurrent model with 1 RCPB is matched with a feedforward model with 2 CPBs, with the same activation functions, loss, etc. 
Fig.~\ref{fig:results:tang:scatter_r_vs_ff:3rd_nips}A also shows the prediction performances of these models when only 50\% and 25\% of the training sets were used, revealing the percentage improvement is more significant when training data is more limited.  
We perform pairwise T-test to evaluate the significance of prediction performance gain for each model type for different amount of the training data, and found the performance gain is significantly greater for the {\bf WF-L}-trained models (performance gain is 11\%, 6\% and 5\% for 25\%, 50\% and 100\% training data) than the {\bf WF-S}-trained models (performance gain is 6.5\%, 3.5\% and 2\% for 25\%, 50\% and 100\% training data). The performance gain for  {\bf RF-S} trained model is significant but small. In short, everything being equal, the 1-RCPB  models and the 2-RCPB models performs consistently better than their 2-CPB and 4-CPB feedforward counterparts respectively, except for the 1-RCPB models trained for the {\bf RF-S} data. For completeness, we showed all 24 panels for each of the three datasets in Appendix~\ref{appendix:ff_model_vs_r}. 
Across all three datasets, we observe that the \texttt{2-avg} (red curve) and \texttt{late-avg} (green curve) readout mode models consistently increase for models of more iterations, and outperform the other models for higher iteration numbers. We suggest this is likely because the temporal averaging of the neural response is explicitly incorporated in these readout modes. 

\subsection{Multi-path approximations work and reveal varying effects of paths of different lengths} 
\begin{figure}[htb]
    \centering
    \includegraphics[width=\textwidth]{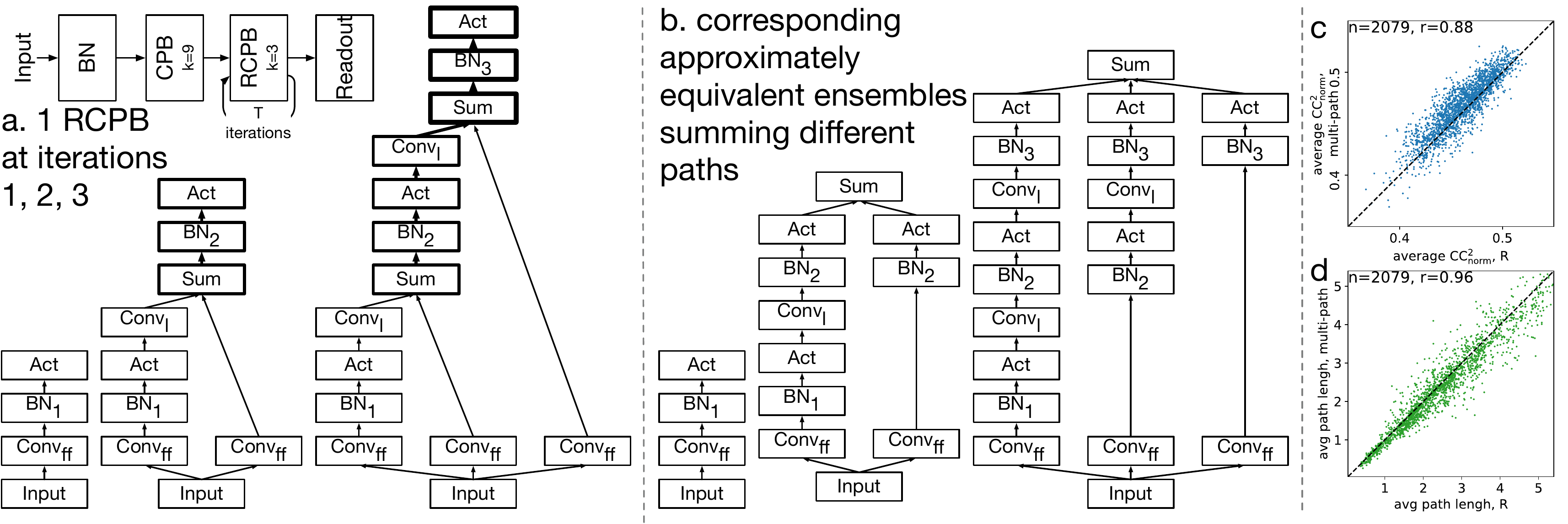}
    \caption{Recurrent computation approximately understood as summation of feed-forward chains of different lengths. \textbf{(a)} shows the information flows of a 1-layer RCPB at iterations 1, 2, and 3 during model inference; the width of a black line is proportional to the number of information flows passing between adjacent components and the width of a component's border is proportional to the the number of information flows passing through. \textbf{(b)} shows the information flows of corresponding multi-path ensembles and each flow sums over feed-forward chains of different depths. \textbf{(c)} compares multi-path models and original recurrent models in terms of model performance. \textbf{(d)} compares the two in terms of average path length of their multi-path ensembles. Models shown here were trained on all the training data. See Appendix~\ref{appendix:model_approximation_multi_path} for details and additional results.}
    \label{fig:results:depth_analysis:schema}
\end{figure}

To verify the approximation of the multi-path ensemble to the original recurrent models,  we compare their neural response prediction performance. Figure~\ref{fig:results:depth_analysis:schema}c shows that their performance metrics are highly correlated across hyperparameter values (r=0.88). Furthermore, the effective average path lengths of the recurrent models and that of the multi-path reformulation are  also highly correlated (r=0.96). This gives us confidence that the approximation is good and so we can draw conclusions about the corresponding recurrent models by analyzing the multi-path ensembles.

\begin{figure}[htb]
     \centering
     \includegraphics[width=\textwidth]{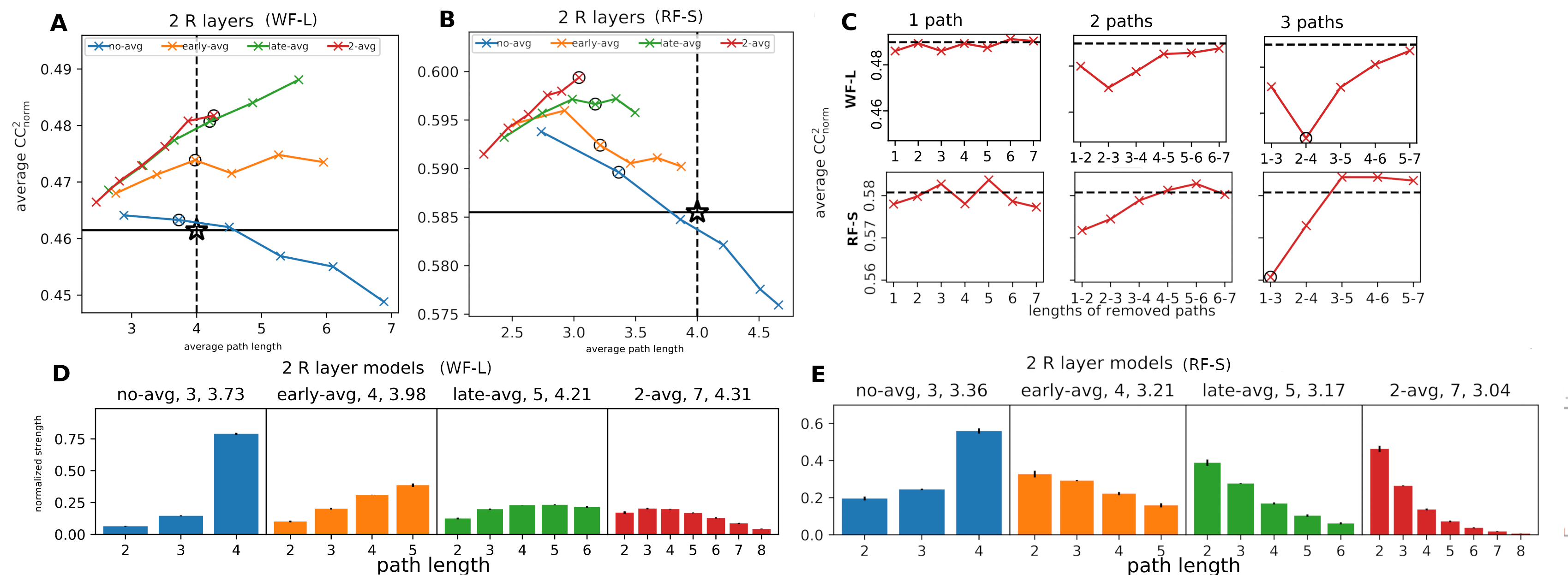} 
     \caption{Path length and diversity across readout modes and datasets. Panels {\bf A} and {\bf B} show average performance as a function of path length for 2-layer recurrent models on the {\bf WF-L} and {\bf RF-S} datasets, respectively. {\bf C} shows the results of ablating paths of one, two, or three lengths for \texttt{models} trained on {\bf WF-L} (top) and {\bf RF-S} (bottom), with shorter paths mattering more for the {\bf RF-S} data (additional results in Appendix~\ref{appendix:ablation}). Panels {\bf D} and {\bf E} show distributions of path lengths for the best-performing iteration numbers of each readout mode on the {\bf WF-L} and {\bf RF-S} datasets respectively. The distributions are more biased towards shorter paths for the {\bf RF-S} models, though in all cases \texttt{no-avg} has a strong bias towards the longest path. 
     }
     \label{fig:results:tang:key_figure}
\end{figure}

The multi-path ensemble reformulation allows us to analyze and dissect the recurrent circuits from the perspective of path-lengths and path distributions. 
Figure~\ref{fig:results:tang:key_figure} panels A and B recast the performance graphs of Figure~\ref{fig:results:tang:scatter_r_vs_ff:3rd_nips}B as a function of effective path lengths instead of iteration numbers for the {\bf WF-L} and {\bf RF-S} datasets ({\bf WF-S} results are somewhere in between and shown in Appendix~\ref{nips2021:appendix:8k_distributions}.  The graph showed that at approximately the same averaged lengths, as indicated by the circled data points, the two readout modes that explicitly models the temporal averaging of the responses at the end, fare better than the 
\texttt{early-avg}, and far better than the \texttt{no-avg} models. 


Given that the notion of diversity is important for ensemble computing,  we investigate  whether the distribution of the contribution of paths of different lengths, beyond just the average length, plays a role in the differential performance of the four readout modes. Figure~\ref{fig:results:tang:key_figure} panels D and E show the distribution of contribution by the paths of different lengths for each readout mode at similar averaged path lengths (the circled points in Figure~\ref{fig:results:tang:key_figure}A and B). We found that the better performing model architecture  tend to be more evenly distributed in path-lengths, with a stronger emphasis on the earlier iterations particularly visible for the {\bf RF-S} dataset. The contribution of the longer paths becomes more significant with the increase in the spatial context of the stimuli and in the duration of the stimulation, which would allow a greater amount of contextual modulation. This finding is also supported by the ablation studies in panel C, the performance of multi-path models retrained with paths of one, two or three consecutive lengths removed (since removing just one length has a very small effect). 
For the {\bf RF-S} dataset, ablating the very shortest paths produces the largest performance decrease, while slightly longer paths (i.e. lengths 2 and 3)  are more important for the {\bf WF-L} trained models. 
Results for models trained on subsets of the {\bf WF} datasets containing only V1 neurons, and the full {\bf WF-S} data, in Appendix~\ref{nips2021:appendix:8k_distributions} and~\ref{nips2021:appendix:multipath_v1_only}, lie somewhat in-between, suggesting that spatial and temporal context are important contributors to the models of neural response.


\subsection{Response dynamics and contextual modulation are most realistic for \texttt{2-avg} models trained on long-presentation data}
\label{results:neurophys}
\begin{figure}[htb]
     \centering
  \includegraphics[width=5.3in]{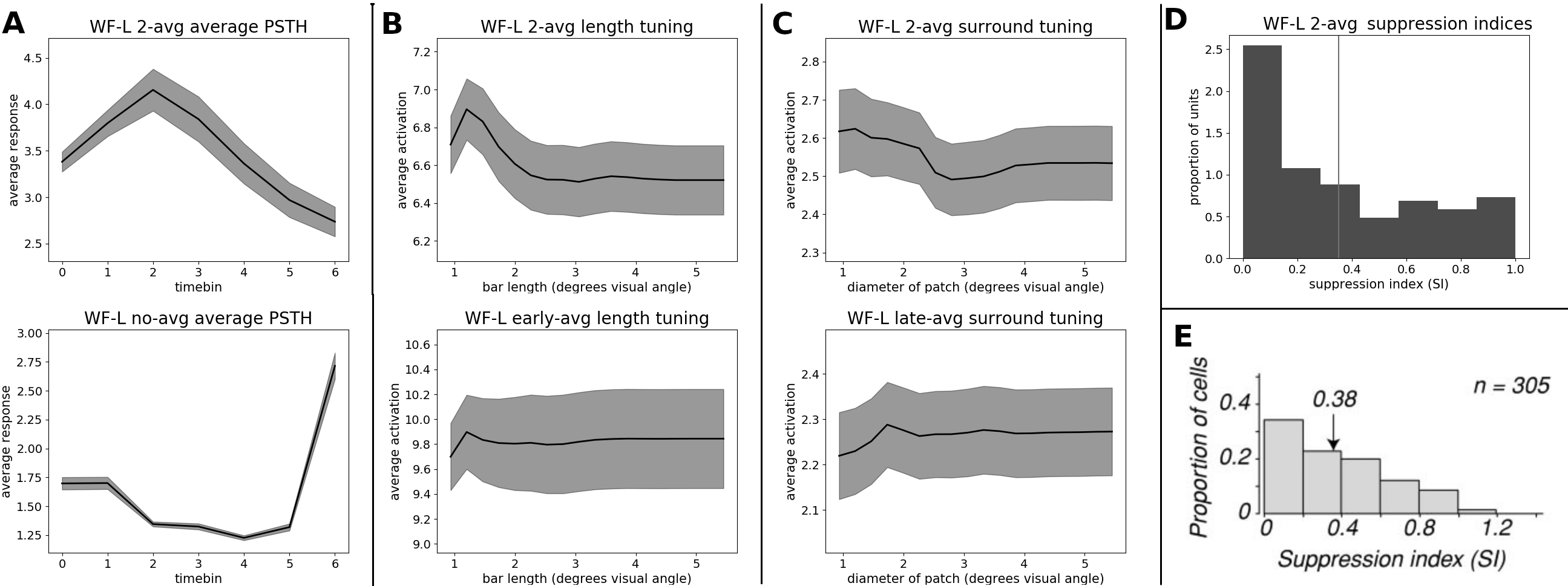}
     \caption{Results of the {\bf WF-L} models on select neurophysiological experiments. The top row shows the responses of units from the RCPB of 1-layer, 7-iteration \texttt{2-avg} models across time ({\bf A}), to bars of increasing length ({\bf B}), and to sine-wave gratings of increasing diameter ({\bf C}).
     Values are averaged over all units from multiple models that showed sustained responses to the stimuli; shaded bands represent $\pm 1$ standard error. The bottom row shows the responses of units from \texttt{no-avg} ({\bf A}), \texttt{early-avg} ({\bf B}), and \texttt{late-avg} ({\bf C}) models in the experiments for their corresponding columns. Each of these readout modes failed to show responses qualitatively matching known V1 response properties in the displayed experiment. Panels {\bf D} and {\bf E} show the distribution and average value of surround modulation indices across neurons for our models' units ({\bf D}) and real V1 neurons ({\bf E}, from~\cite{cavanaugh2002}).} 
     \label{neurophys_fig}
\end{figure}

Figure~\ref{neurophys_fig} contains results from the contextual modulation experiments described in Section~\ref{methods:neurophys} for 7-iteration models of the different readout modes trained on the {\bf WF-L} data. 
The \texttt{2-avg} models were the only ones to exhibit all three tested phenomena; the failed results for each of the other modes are shown in the bottom row of panels A, B, and C. In addition, the feedforward component of the \texttt{2-avg} models, as well as those trained on the {\bf RF-S} data, fail to exhibit some phenomena (Appendix~\ref{appendix:neurophys}). Figure~\ref{neurophys_fig}A shows the averaged temporal response to sine-wave grating stimuli; the initial burst followed by a decay to a steady state of the \texttt{2-avg} matches visual cortical neurons~\citep{Zipser7376}.
Two well-observed physiological effects that have been thought to arise from recurrent circuits are end-stopping and surround suppression. End-stopping is the reduction of neural responses to a bar of the optimal orientation of the neuron when the bar's length exceeds the size of the receptive field or is longer than a preferred length~\citep{doi:10.1152/jn.1965.28.2.229}. Surround suppression is the reduction of neural responses when the size of the sine-wave grating of a neuron's preferred orientation is larger than the receptive field, encroaching on the receptive field's near and/or far surround~\citep{doi:10.1146/annurev.ne.08.030185.002203}. 
Figure~\ref{neurophys_fig}B and C display the average tuning of model units to bars and gratings as a function of length/size, showing the expected rise-and-fall for \texttt{2-avg} units, but not for \texttt{early-avg} and \texttt{late-avg} respectively. 
In addition, the average RF size of about 1.2 degrees of visual angle (\~9 pixels in the models' input) is roughly the expected RF size of V1 neurons at our eccentricity, and the average level of surround modulation, compared to that of real neurons in~\cite{cavanaugh2002}, is quite close (Figure~\ref{neurophys_fig}D and E). 

%% file: discussion_new.tex
\section{Discussion}
\label{nips2021:discussion_new}


In this study, we showed that recurrent models with the appropriate computational architecture significantly outperform feedforward models of matched size and hyperparameters in scenarios where  neural responses are significantly modulated by the spatial and temporal context.  The performance gain is significant and consistent, particularly when the readout mode takes into account the experimenter's temporal average of the neural responses in constructing the target responses. 
Furthermore, it is reassuring that the hidden units in \texttt{2-avg} models
 exhibit similar temporal dynamics and contextual modulation as observed in early visual cortical neurons in neurophysiological experiments, even though the models are trained only to predict mean responses with no constraints imposed explicitly to bring about these effects. 
Taken together, these results suggest that neural response prediction can provide a useful paradigm for characterizing the underlying recurrent circuits of the visual cortex. 

By conceptualizing and reformulating recurrent networks as multi-path ensembles, we suggest  recurrent circuits might derive their computational advantages from the notion of ensemble computing. More importantly, the reformulation allows us to perform analysis of the contribution of different paths based on  path-length statistics and ablation analysis. The results of our analysis suggest models with a more uniform distribution of path-lengths and paths of longer lengths are needed to model greater amount of contextual modulation. The recognition that recurrent circuits can be conceptualized as multi-path ensembles might suggest a new direction of research to investigate how more powerful multi-path ensemble can be constructed and optimized by using recurrent network, both for machine learning and for understanding the brain.


There are several limitations to our work. While using small models is standard for neural response prediction, and allows us to explore a wide variety of hyperparameter combinations, the method of multi-path reformulation developed for such shallow models might be difficult to transfer to the much deeper models generally used in computer vision, or to other domains in which recurrent networks are applied. This means that the positive societal benefits of our work resulting from greater model interpretability may be limited, but we do not foresee any negative impacts from this formulation or better characterization of early visual cortical circuits. The hyperparameters we sweep over may not include those that result in optimal performance, and architectural improvements like the Gaussian readout of~\cite{lurz2021generalization} may reduce the improvement from recurrent connections that we find. In addition, our recurrent circuit models are far simpler than the true circuits in visual cortex, which include a detailed laminar structure, extensive feedback, dynamically varying connectivity, and more. Thus, there is a limit to how well neural circuits can be characterized by our current models.

%% file: appendix.tex
\section{Details of neural data sets}
\label{appendix:neural_data}







The {\bf WF-S} and {\bf WF-L} datasets were collected in our laboratory from  an awake behaving macaque monkey performing fixation task using an SC96 array, a 96-electrode multi-electrode array (Gray Matter Research, MT), implanted over the V1 operculum.
Each set was obtained from 5 days of recording, with 4000 trials per day. 

The {\bf RF-S} is the dataset from Cadena et al. 2019 Plos Computational Biology, collected using 32-channel linear multi-electrode array. The data are available publicly (https://doi.gin.g-node.org/10.12751/g-node.2e31e3/) under a Creative Commons license, obtained and used in this paper with permission from the owners.

\begin{figure}[htb]
     \centering
     \includegraphics[width=0.75\textwidth]{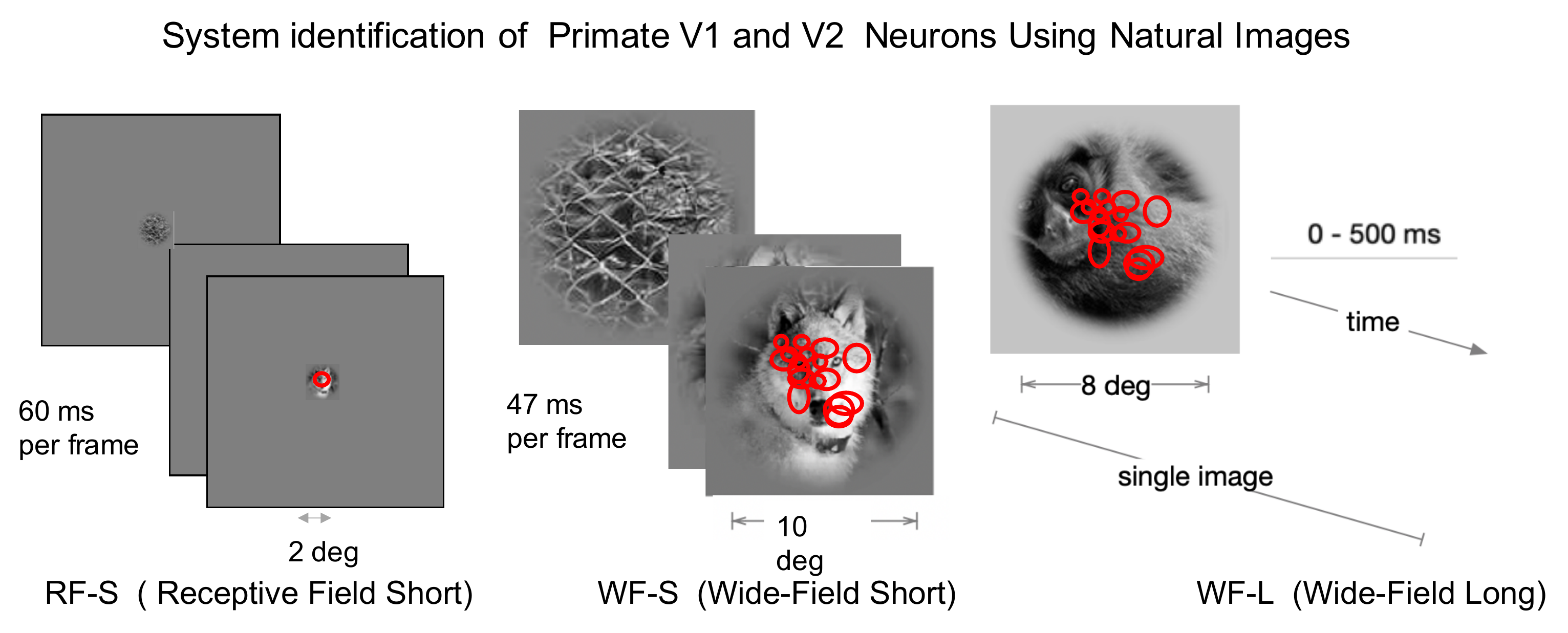}
    \caption{Stimulus presentation paradigms of a single trial. Stimuli (8000 and 7250 respectively) were presented in rapid sequence with each stimulus presented for 47 or 40 ms respectively for {\bf WF-S} and {\bf RF-S}. Each stimulus was presented for 500 ms per trial for {\bf WF-L}, and responses to 2250 images (8-10 repeats each) were collected across five days. Only neurons that could be tracked and exhibit consistent tuning to daily calibration images across those days were kept for analysis.}
    \label{fig:methods:paradigm}
\end{figure}

\section{Details of the explored models}
\label{appendix:model_math}

\subsection{Baseline models}
\label{appendix:model_math:baseline}

During inference, a baseline model takes an image $\mat{x} \in \mathbb{R}^{H \times W}$ of height $H$ and width $W$ as input and generates an prediction $\vect{\hat{r}} \in \mathbb{R}^{N}$ of the neural responses $\vect{r} \in \mathbb{R}^{N}$ to the image. Mathematically, the model inference process is defined by Eqs.~\eqref{eqs:datadriven-feedforward-model}.

\begin{subequations}
\label{eqs:datadriven-feedforward-model}
\begin{align}
    \vect{\hat{r}} &= \nnact(\nnfc(\nnavgpool(\mat{y}^{(M)}))) \label{eqs:datadriven-feedforward-model:final} \\
    \mat{y}^{(m)} &= \nncpb^{(m)}(\mat{y}^{(m-1)}) \quad m=2,\ldots,M \label{eqs:datadriven-feedforward-model:later} \\
    \mat{y}^{(1)} &= \nncpb^{(1)}(\nnbn(\mat{x})) \label{eqs:datadriven-feedforward-model:start} \\
    \nncpb^{(m)}(\mat{y}) &= \nnact(\nnbn^{(m)}(\nnconv^{(m)}(\mat{y}))) \label{eqs:datadriven-feedforward-model-cpb}
\end{align}
\end{subequations}

For a model of $M$ convolutional processing blocks (CPBs), the model inference starts with Eq.~\eqref{eqs:datadriven-feedforward-model:start} to normalize the input image and obtain the initial CPB's output $\mat{y}^{(1)}$, followed by a few applications of Eq.~\eqref{eqs:datadriven-feedforward-model:later} to get outputs of later CPBs $\mat{y}^{(2)},\ldots \mat{y}^{(M)}$. Finally, Eq.~\eqref{eqs:datadriven-feedforward-model:start} is applied to obtain the predicted neural responses $\vect{\hat{r}}$. $\nnconv$, $\nnbn$, $\nnact$, $\nnavgpool$, $\nnfc$ represent convolution, batch normalization, nonlinear activation, average pooling, and factorized fully connected layers \citep{DBLP:conf/nips/KlindtEEB17}, respectively.  Different convolutional processing blocks have different hyper parameters and learned parameters, as denoted by superscripts $(m)$ in Eq.~\eqref{eqs:datadriven-feedforward-model-cpb}. 

\subsection{Recurrent models}
\label{appendix:model_math:new}

\begin{figure}[htb]
    \centering
    \includegraphics[width=0.75\textwidth]{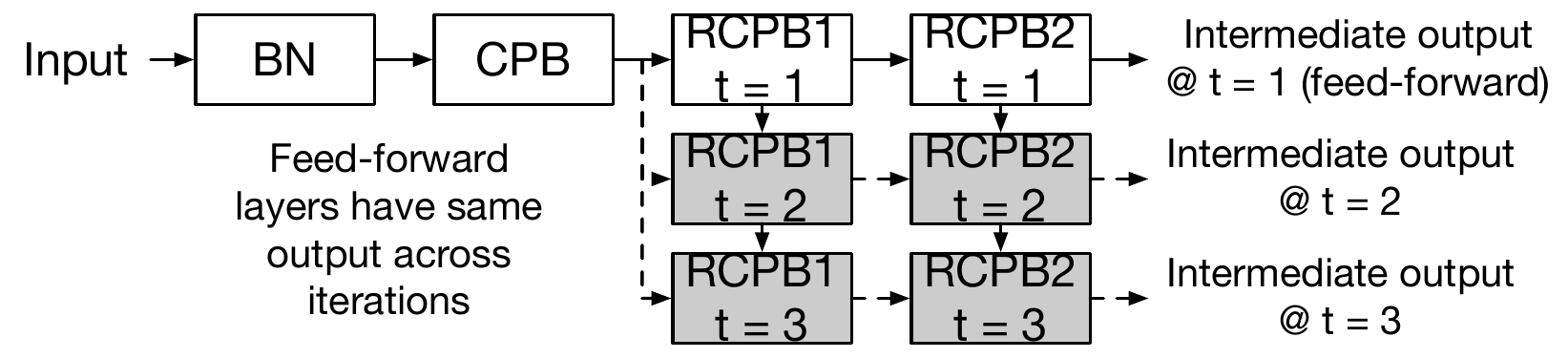}
    \caption{Information flow in the convolutional phase for a recurrent model. The example model has one CPB, two RCPBs, and three iterations in total. In the first iteration $t=1$, the information flow is the same as that in a baseline feed-forward model. In later iterations, the output of a RCPB depends on both bottom-up and lateral inputs. The extra information flow of a recurrent model relative to a feed-forward one is shown in shaded blocks and dashed lines.}
    \label{fig:methods:datadriven_k_models_unroll}
\end{figure}

The model inference process is defined by Eqs.~\eqref{eqs:datadriven-recurrent-model}. and the detailed information flow for a recurrent model across iterations are illustrated in Figure~\ref{fig:methods:datadriven_k_models_unroll}.

\begin{subequations}
\label{eqs:datadriven-recurrent-model}
\begin{align} 
    \vect{\hat{r}}_{\texttt{no-avg}} &= \nnact(\nnfc(\nnavgpool(\mat{y}^{(M,T)}))) \label{eqs:datadriven-recurrent-model:inst-last} \\
    \vect{\hat{r}}_{\texttt{early-avg}} &= \nnact(\nnfc( \nnavgpool(\mat{\overline{y}}^{(M,T)} ))) \label{eqs:datadriven-recurrent-model:cm-last} \\
    \vect{\hat{r}}_{\texttt{late-avg}} &= \frac{1}{T} \sum_{t=1}^T \nnact(\nnfc(\nnavgpool( \mat{y}^{(M,t)} ))) \label{eqs:datadriven-recurrent-model:inst-avg} \\
    \vect{\hat{r}}_{\texttt{2-avg}} &= \frac{1}{T} \sum_{t=1}^T \nnact(\nnfc(\nnavgpool( \mat{\overline{y}}^{(M,t)} ))) \label{eqs:datadriven-recurrent-model:cm-avg} \\
    \mat{\overline{y}}^{(M,t)} &= \frac{1}{t} \sum_{t'=1}^t \mat{y}^{(M,t')} \\
    \mat{y}^{(m,t)} &= \nnrcpb^{(m,t)}(\mat{y}^{(m-1,t)},\mat{y}^{(m,t-1)}) \quad m=2,\ldots M; t=1,\ldots T \label{eqs:datadriven-recurrent-model:inst-later} \\
    \mat{y}^{(1,t)} &= \nnrcpb^{(1,t)}(\nnbn(\mat{x}),\mat{y}^{(1,t-1)}) \quad t=1,\ldots T \label{eqs:datadriven-recurrent-model:inst-init} \\
    \mat{y}^{(m,0)} &= \mat{0} \\
    \nnrcpb^{(m,t)}(\mat{y}, \mat{y}') &= \nnact(\nnbn^{(m,t)}(\nnconv^{(m)}_{\text{feed-forward}}(\mat{y}) + \nnconv^{(m)}_{\text{lateral}}(\mat{y}'))) \label{eqs:datadriven-recurrent-model-rcpb}
\end{align}
\end{subequations}

For a model of $M$ recurrent convolutional blocks (CPBs) and $T$ iterations in total, the model inference starts with $T$ cycles of Eqs.~\eqref{eqs:datadriven-recurrent-model:inst-init},\eqref{eqs:datadriven-recurrent-model:inst-later} to obtain the responses of all $M$ RCPBs across $T$ iterations $\mat{y}^{(m,t)}, m=1\ldots M, t=1\ldots T$. Finally, one of Eqs.~\eqref{eqs:datadriven-recurrent-model:inst-last},\eqref{eqs:datadriven-recurrent-model:cm-last},\eqref{eqs:datadriven-recurrent-model:inst-avg},\eqref{eqs:datadriven-recurrent-model:cm-avg} is used to obtain the final model output depending on the readout mode used. The readout modes model the computations performed by or on the target neurons to generate the measured neural responses (see main manuscript for motivation and discussion). There are two possible stages of averaging. The first stage (early stage) that integrates the intermediate outputs of the hidden units in the recurrent across iterations. The second stage (late stage) average the temporal response of the target neurons to simulate the computation we performed on the neurons' PSTH. 

\begin{enumerate}
    \item \texttt{no-avg} (Figure~\ref{fig:methods:readout}a)  No averaging. The intermediate output at the last iteration is passed through the remaining layers of the model to get the final output.
    \item \texttt{early-avg} (Figure~\ref{fig:methods:readout}b)  Early averaging only. It's similar to \texttt{no-avg} except that the average of intermediate outputs across iterations is passed through the remaining layers of the model.
    \item \texttt{late-avg} (Figure~\ref{fig:methods:readout}c)  Late averaging only. The neural response predictions based on the individual intermediate outputs across iterations are averaged as the final output.
    \item \texttt{2-avg} (Figure~\ref{fig:methods:readout}d)  Both early and late averaging. First the cumulative averages of intermediate outputs across iterations are computed and then the neural response predictions based on these cumulative averages are further averaged as the final output.
\end{enumerate}

\section{Neurophysiological experiments}
\label{appendix:neurophys}
We tested the hidden units in the recurrent layer block in our models using  oriented bars and sine-wave gratings, as commonly used in visual neuroscience, to evaluate whether these units exhibit  end-stopping and surround suppression effects, even though the models were not designed or trained for these effects.  We extracted the responses of units in the center position of the RCPB, so their receptive fields were located near the center of the image.
\subsection{End-stopping}
First, we determined the precise RF location and preferred orientation of each unit with oriented bars of an intermediate length (10 pixels) presented in a $5 \times 5$ grid of locations around the center of the image, spaced 5 pixels apart. The bars spanned the range of 0 to 360 degrees of orientation, equally spaced at 22.5 degree intervals, for a total of 16 orientations, and were all 2 pixels wide, black against a white background. Using the orientation and position of bar that maximized the response of each unit in the first iteration, we presented bars varying in length from 5 to 45 pixels, at intervals of 2 pixels. For the analysis in Figure 6 of the main paper, we included only units that had a stable response of at least 1.0 across all time bins to at least one of the lengths, since some units were observed to drop to 0 or very low values. About 60-80\% of units across models had these stable responses.

\subsection{Surround suppression}
We determined the orientation tuning of units to sine-wave gratings separately from that to oriented bars, using gratings that covered the whole image. However, we used the receptive field locations determined by responses to oriented bars for this experiment. Gratings varied in frequency from 0 to 15 cycles per image (roughly the maximum achievable with the low-resolution images the models were trained on) at steps of 1 Hz, and the optimal frequency was selected for each unit. Then, gratings of increasing diameter, from 5 to 45 pixels stepping by 2, were presented at each unit's preferred location and frequency. The gratings were presented on a gray background, with slight Gaussian blurring at the end of the aperture to reduce edge effects, and at 30\% contrast, since surround suppression is generally observed at lower contrasts, and this value resulted in the closest average suppression index to previously reported neurophysiological data. 
Like for end-stopping, we only included units with a stable response to at least one of the grating sizes, which included about 60-80\% of units.

\subsection{Temporal dynamics}
The average temporal dynamics were computed over all units' responses to all orientations and frequencies of the full-image sine-wave gratings used to determine orientation tuning for the surround suppression experiment. Dynamics in response to oriented bars looked the same. 

\subsection{Feedforward responses and RF-S trained \texttt{2-avg} models do not exhibit V1-like responses}

\begin{figure}[htb]
    \centering
    \includegraphics[width=0.55\textwidth]{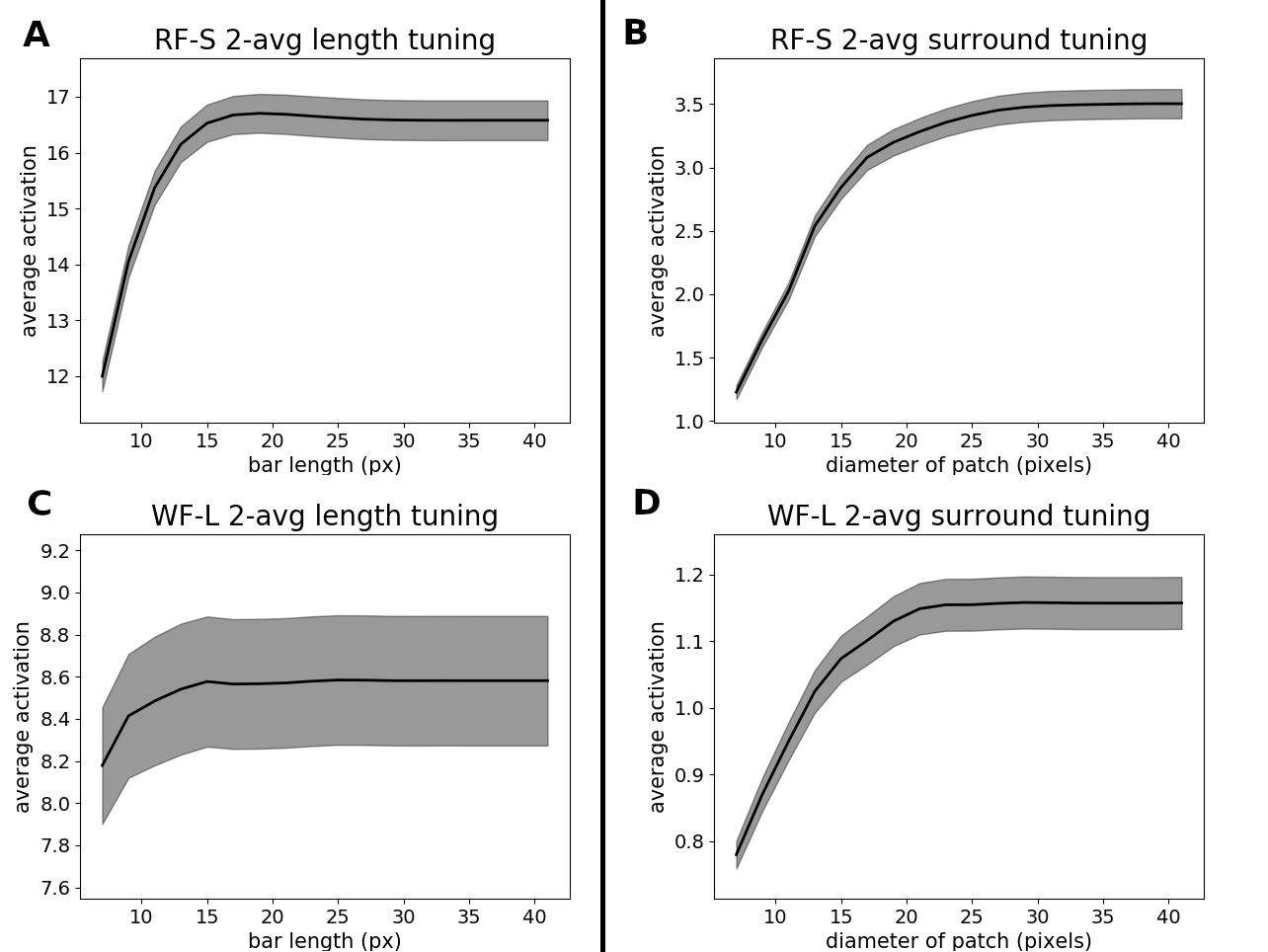}
    \caption{Results on the end-stopping and surround-suppression neurophysiological experiments for the \texttt{2-avg} models in circumstances where they fail to reproduce V1-like effects. Panels {\bf A} and {\bf B} show the results for \texttt{2-avg} models trained on the {\bf RF-S} data instead of the {\bf WF-L} data of the main paper's Figure 6, using the last iteration as in the main paper, for end-stopping and surround suppression respectively. Panels {\bf C} and {\bf D} show the results for \texttt{2-avg} models trained onthe {\bf WF-L} data, but using the very first iteration of the response, before recurrent connections have come into play. The structure of the plots is as described in the caption of Figure 5 in the main paper. In all cases, end-stopping and surround suppression are not present.}
    \label{fig:neurophys_appendix}
\end{figure}

Figure~\ref{fig:neurophys_appendix} shows the results of \texttt{2-avg} models trained on the {\bf RF-S} data on our experiments (panels {\bf A} and {\bf B}), as well as the feedforward responses (first iteration) of models trained on the {\bf WF-L} data. In all of these cases, end-stopping and surround suppression are not shown. The lack of appropriate contextual modulation in the first iteration of the response shows that the effects arise as a result of the learned recurrent circuit, and are not just a property of the feedforward part of the RCPBs. The lack of modulation in the {\bf RF-S}-trained models suggests that neural responses exhibiting surround modulation (like the long-presentation, wide-field {\bf WF-L} data) are necessary to learn the appropriate circuits via neural response prediction.

\subsection{2-recurrent-layer WF-L trained models exhibit similar responses as 1-layer ones}

\begin{figure}[htb]
    \centering
    \includegraphics[width=0.9\textwidth]{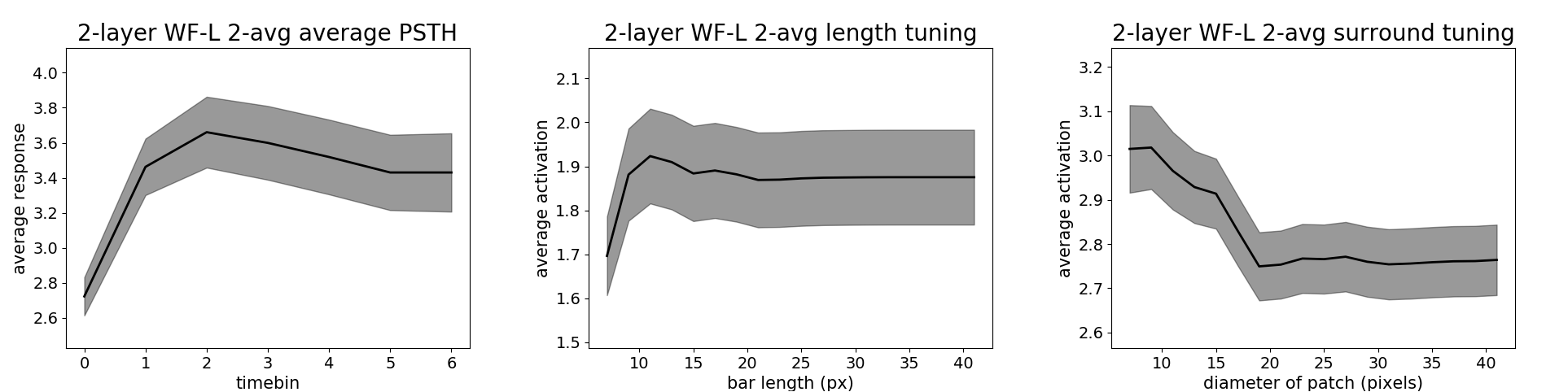}
    \caption{Figures corresponding to the top row of Figure 6 of the main paper, but for models with two recurrent layers. The hidden units used are the ones in the second recurrent layer. Similar temporal dynamics, end-stopping, and surround suppression obtain.}
    \label{fig:2layer_neurophys_appendix}
\end{figure}

Figure~\ref{fig:2layer_neurophys_appendix} shows the responses of hidden units from the second recurrent layer of two-recurrent-layer models on our experiments. The temporal dynamics, end-stopping, and surround suppression are still present for {\bf WF-L}-trained models, in their last iteration after recurrent processsing.

\section{Implementation details}
\label{appendix:implementation_details_all}

\subsection{Data preprocessing}
\label{method:implementation-details:datapreprocessing}

Following practices in the literature \citep{DBLP:conf/nips/KlindtEEB17,10.1371/journal.pcbi.1006897}, for each data set, we used cropped and downsampled images as the input and normalized average neural responses as the output for modeling purposes.

\subsection{Model configurations}
\label{nips2021:method:implementation-details:hyperparameters:all}
To comprehensively evaluate recurrent models vs.\@ feed-forward models, we trained and tested multiple variants of recurrent and feed-forward models with different model configurations.



\paragraph{Kernel sizes}

Following previous studies \citep{10.1371/journal.pcbi.1006897,DBLP:conf/nips/KlindtEEB17} and as well as  based on our pilot experiments, in either of a feed-forward model or a recurrent model, the first convolutional layer has a relatively large kernel size (\num{9}) and subsequent layers have a small kernel size of \num{3}. The pooling layer always has a kernel size of \num{3} and a stride of \num{3}.

\paragraph{Feed-forward models} For each data set, we trained  a large set of feed-forward models with different configurations differing on aspects like amount of training data, model size, loss functions, randomness parameters, etc., as listed in Table~\ref{tab:hyperparameters-feedforward}.  Each aspect of configuration was explored for certain reasons as follows.

\begin{itemize}
    \item \textbf{Training data} and \textbf{model size-related model hyperparameters}. we explored model configurations in these two aspects because we feel training data amount and model size affect the relative performance difference between feed-forward and recurrent models. In particular, we hypothesized that the additional circuit priors from recurrent models would be more useful when there was less training data and the model size was larger.
    \item \textbf{Model size-independent model hyperparameters and loss functions} We explored model configurations in this aspect because we wanted to know if the advantage of recurrent models over feed-forward models is limited to certain choices of model size-independent model hyperparameters and loss functions, or if such advantage is universal across choices of model size-independent hyperparameters and loss functions.
    \item \textbf{Randomness hyperparameters} We explored different model initialization seeds because we wanted to obtain robust metric numbers for our models by averaging results over different seeds. Ideally we should also explore over other randomness hyperparameters, such as the seed used to split data into training, validation, and testing sets. We did not do so due to computational resource constraints.
\end{itemize}

\begin{table}[htb]
\caption{Model configurations explored for feed-forward models. k denotes the kernel size for a convolutional block}
\label{tab:hyperparameters-feedforward}
\begin{tabular}{@{}lll@{}}
\toprule
Aspect & name             & values explored                       \\ \midrule
\begin{tabular}[c]{@{}l@{}}training data\end{tabular} &
  amount of training data &
  \begin{tabular}[c]{@{}l@{}}25\%, 50\%, 100\%\\(1280, 2560, 5120 for {\bf WF-S},\\
  1160, 2320, 4640 for {\bf RF-S},\\
  350, 700, 1400 for {\bf WF-L})\end{tabular} \\ \midrule
\begin{tabular}[c]{@{}l@{}}model size-related\\ model hyperparameters\end{tabular} &
  \# of convolutional blocks & \begin{tabular}[c]{@{}l@{}}3 (1 CPB \makeatletter{}@\makeatother{} k=9 + 2 CPBs \makeatletter{}@\makeatother{} k=3),\\5 (1 CPB \makeatletter{}@\makeatother{} k=9 + 4 CPBs \makeatletter @\makeatother{} k=3)\end{tabular} \\ \cmidrule(l){2-3} 
 &
  \begin{tabular}[c]{@{}l@{}}\# of channels per\\ convolutional layer\end{tabular} &
  \begin{tabular}[c]{@{}l@{}}8, 16, 32, 48, 64 for {\bf WF-S, RF-S},\\ 8, 16, 32 for {\bf WF-L} \end{tabular} \\ \midrule
\begin{tabular}[c]{@{}l@{}}model size-independent\\ model hyperparameters\\ and loss functions\end{tabular} & loss function    & mean squared error, mean Poisson loss \\ \cmidrule(l){2-3} 
         & activation layer & ReLU, softplus                        \\ \cmidrule(l){2-3} 
 &
  \begin{tabular}[c]{@{}l@{}}order of BN and activation\\ in the first convolutional block\end{tabular} &
  BN before act, BN after act \\  \midrule
randomness hyperparameters & 
  model initialization seed &
  0, 1 \\ \bottomrule
\end{tabular}
\end{table}

\paragraph{Recurrent models}
\label{nips2021:method:implementation-details:hyperparameters:recurrent}
For each feed-forward model, we trained \num{24} corresponding recurrent models of the same size, to comprehensively compare feed-forward and recurrent models. To keep the comparison as fair as possible, each of the \num{24} recurrent models have the same hyperparameters as the feed-forward one on those hyperparameters listed in Table~\ref{tab:hyperparameters-feedforward}, except with a decreased number of convolutional layers to match the model size because a recurrent convolutional processing block has roughly twice as many parameters as a convolutional processing block with the same number of channels. In particular, we replace every two CPBs of kernel size 3 in a feed-forward model with one RCPB of kernel size 3 in a recurrent model. Note that a recurrent model does not involve recurrent computation in its first CPB, which has a large kernel size of 9, and only has recurrent computation at subsequent CPBs of kernel size 3; we made this decision based our pilot experiments and for simplicity in matching the model size between recurrent and feed-forward models. In addition, the \num{24} recurrent models differ in their recurrence-related model hyperparameters listed as follows.

\begin{itemize}
    \item \textbf{Number of iterations}, with six possible values \numrange{2}{7}. This hyperparameter affects the amount of recurrent computation during model training and inference. We wanted to explore this hyperparameter because obviously amount of recurrence may affect model performance. Note that we always use the same number of iterations during training and testing; e.g.\@ a recurrent model with $T$ iterations during training is always evaluated with $T$ iterations during testing.
    \item \textbf{Readout mode}, with four possible values (\texttt{no-avg}, \texttt{early-avg}, \texttt{late-avg}, \texttt{2-avg}. We want to explore this hyperparameter because we wanted to know if some readout modes perform better than others and if there exists some optimal way to make use of intermediate outputs generated by a recurrent model.
\end{itemize}

\subsection{Model optimization}
\label{method:implementation-details:optimization}

We trained all models using PyTorch \citep{2019arXiv191201703P} and followed \citet{DBLP:conf/nips/KlindtEEB17} for model optimization.

\paragraph{Objective function}

For each model, the objective function to be minimized was the sum of two parts.

\begin{description}
    \item[Neural prediction loss] Depending on the choice of hyperparameters, the loss can be either mean squared loss or mean Possion loss between between the predicted neural response and recorded ground truth averaged across neurons and images. Note that for recurrent models with \texttt{late-avg} or \texttt{2-avg} readout modes, following the practice in \citet{10.1371/journal.pcbi.1008215}, we computed the neural prediction loss by averaging the losses over individual iterations. 
    \item[Regularization terms] We applied L1 sparsity penalties as regularization; in addition, for the first convolutional layer of the model, we applied smoothness regularization as in Section~5.2 of \citet{DBLP:conf/nips/KlindtEEB17}.
\end{description}

\paragraph{Optimization algorithm}

Given the objective function, we used the same optimization algorithm as in the public code of \citet{DBLP:conf/nips/KlindtEEB17}; in particular model parameters were optimized sequentially in three phases, by three Adam optimizers with decreasing learning rates. Early stopping was applied in each phase, guided by the neural prediction loss evaluated on the validation set. For each data set, roughly \SI{64}{\percent}, \SI{16}{\percent},and \SI{20}{\percent} of images were used for training, validation, and testing, respectively. Due to computation resource constraints, we only created one particular split of training, validation, and testing sets for each dataset. 

\section{Model performance evaluation in detail}
\label{appendix:model_performance}

Given a trained model, we use average $\ccnorm^2$ over all neurons to quantify its performance on a data set. For each neuron, we compute its $\ccnorm^2$ based on Eqs.~\eqref{eqs:ccnorm2}.

\begin{subequations}
\label{eqs:ccnorm2}
\begin{align}
\ccnorm^2 &= \frac{\ccraw^2}{\ccmax^2} \label{eq:ccnorm2_inner} \\
\ccraw &= \operatorname{Pearson}(\vect{r}, \vect{\hat{r}}) \label{eq:eval_ccabs} \\
\ccmax &= \sqrt{\frac{\mathrm{Var}(\{ \sum_k r_{m,k} \})-\sum_k \mathrm{Var}(\{ r_{m,k} \}) }{K(K-1) \mathrm{Var}(\{ r_{m} \}) }}. \label{eq:eval_ccmax} \\
\vect{r} &= (r_1, r_2, \ldots, r_M)  \label{eq:eval_vectr} \\
r_{m} &= \frac{\sum_{k=1}^{K} r_{m,k}}{K}  \label{eq:eval_rm}
\end{align}
\end{subequations}

Concretely, we first compute the raw Pearson correlation $\ccraw$ between the ground truth trial-averaged neural responses $\vect{r}$ as defined by Eqs.~\eqref{eq:eval_vectr},\eqref{eq:eval_rm} and the model responses $\hat{\vect{r}}$ using Eq.~\eqref{eq:eval_ccabs}. We then divide $\ccraw$ by $\ccmax$, which is defined in Eq.~\eqref{eq:eval_ccmax} and estimates the maximal Pearson correlation coefficient an ideal model can achieve given the noise in the neural data \citep{10.3389/fncom.2016.00010,Hsu:2004ku}; inside the square root of Eq.~\eqref{eq:eval_ccmax}, the numerator is the difference between variance of response sums per stimulus $\sum_k r_{m,k}$ and sum of response variances per trial $\mathrm{Var}(\{ r_{m,k} \})$, and the denominator is the variance of the trial-average neural responses scaled by a factor related to the number of trials $K$. Finally, we get the square of the normalized Pearson correlation coefficient using Eq.~\eqref{eq:ccnorm2_inner}. As squared $\ccraw$ gives the fraction of variance in neural responses explained by the model in a simple linear regression, squared $\ccnorm$ gives the normalized explained variance that accounts for noise in the neural data. Note that we used neural responses on all stimuli instead of on testing set stimuli to compute $\ccmax$ in Eq.~\eqref{eq:eval_ccmax} for more accurate estimation. In addition, because the {\bf WF-L} data set has a variable number of trials per image (8 to 10), we only used the first eight trials for simplicity. 

While the derivation of $\ccmax$ is technically complicated, it's positively related with the Pearson correlation between mean responses computed over half of trials and those computed over the other half \citep{10.3389/fncom.2016.00010,Hsu:2004ku}. We use $\ccnorm^2$ instead of raw $\ccraw^2$ to quantify model performance because the former is more robust to trial to trial response fluctuations. Our main results in Section~\ref{results:datadriven_models} still hold if we measure the performance in $\ccraw^2$. 

\FloatBarrier


\section{Path length calculations for the multi-path models}
\label{appendix:model_approximation_multi_path}

\begin{figure}[htb]
    \centering
    \includegraphics[width=0.85\textwidth]{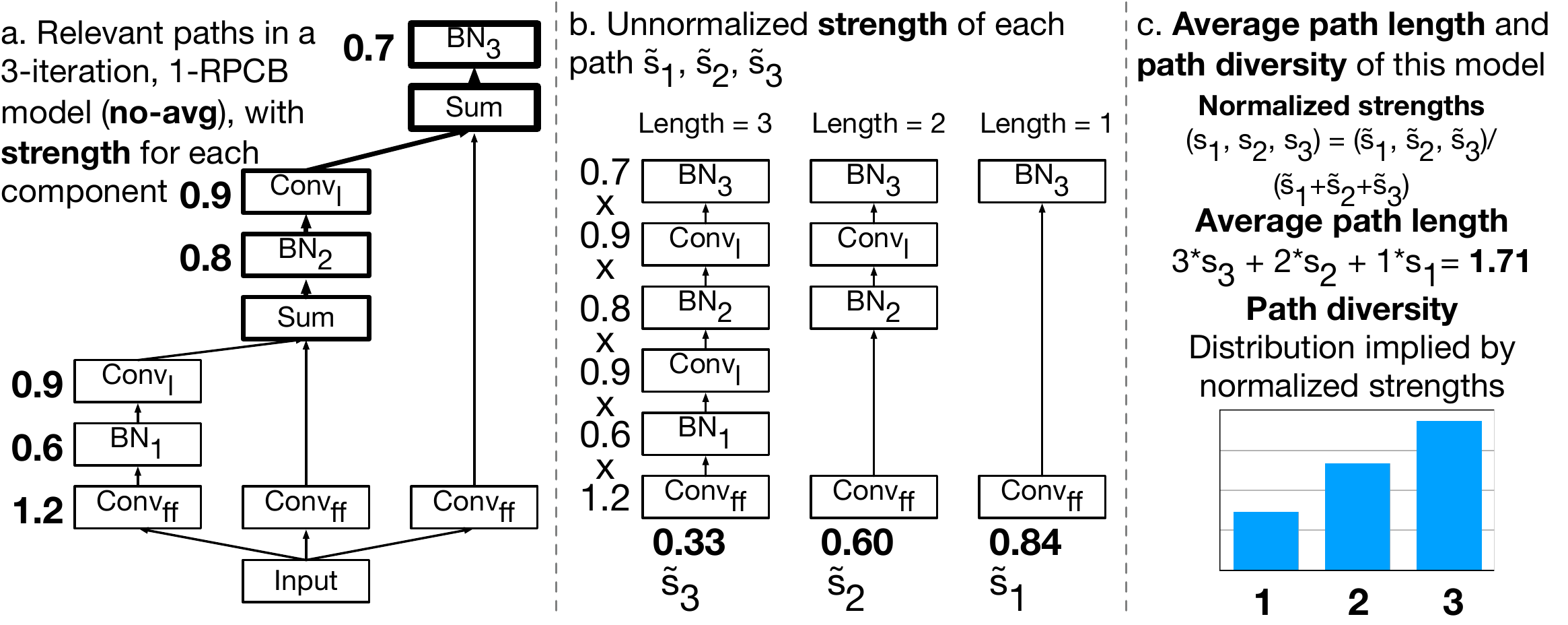}
    \caption{Computation of average path length and path diversity for an example recurrent model. The model has one recurrent block (RCPB) and uses the \texttt{no-avg} readout mode with three iterations. In \textbf{(a)}, the relevant information flow is shown, with a \emph{strength} assigned to each component. In \textbf{(b)}, the original information flow is reformulated as a simple multi-path ensemble with three paths, and the unnormalized strength of each path $\tilde{s}_1,\tilde{s}_2,\tilde{s}_3$ is computed as the product of strengths along that path. In \textbf{(c)}, the \emph{average path length} of the model is defined as the average path length weighted by the normalized strengths $s_1,s_2,s_3$, and the \emph{path diversity} of the model is the discrete probability distribution implied by normalized strengths $s_1,s_2,s_3$. The length of each path is defined as the number of convolutional layers in that path. Activation layers are omitted in \textbf{(a)} and \textbf{(b)} for brevity and they do not affect the calculation due to having a strength of \num{1}.}
    \label{fig:results:depth_analysis:ensemble_characteristic}
\end{figure}

The relative contributions of each path of a multi-path model can be quantified; summarizing these contributions by path length allows us to compute the average path length of each model, as well as its distribution, or diversity, as presented in Figure 5 of the main paper.
Fig.~\ref{fig:results:depth_analysis:ensemble_characteristic} demonstrates the computation of average path length and path diversity for an example recurrent model. The computation of length and diversity depends on the concept of \emph{strength} for each component and each path. For a component on some path, roughly speaking, the strength of the component is a scalar measuring the ratio of the magnitude of the component's output over that of its input. In particular, we define the \emph{strength} of each component as follows: the strength of a BN layer is the average of the absolute values of the scaling factors, over all output channels; the strength of a convolutional layer is the average of the 2-norms of the 3-D convolutional kernels flattened into vectors, over all output channels; the strength of an activation layer (ReLU or softplus in this study) is \num{1}, as the activation layer outputs the input itself when the input is large enough. For a path, we define its strength to be the product of the strengths of all components along the path.

\FloatBarrier

\section{Performance versus iterations divided by number of recurrent layers} 

\begin{figure}[htb]
    \centering
    \includegraphics[width=0.7\textwidth]{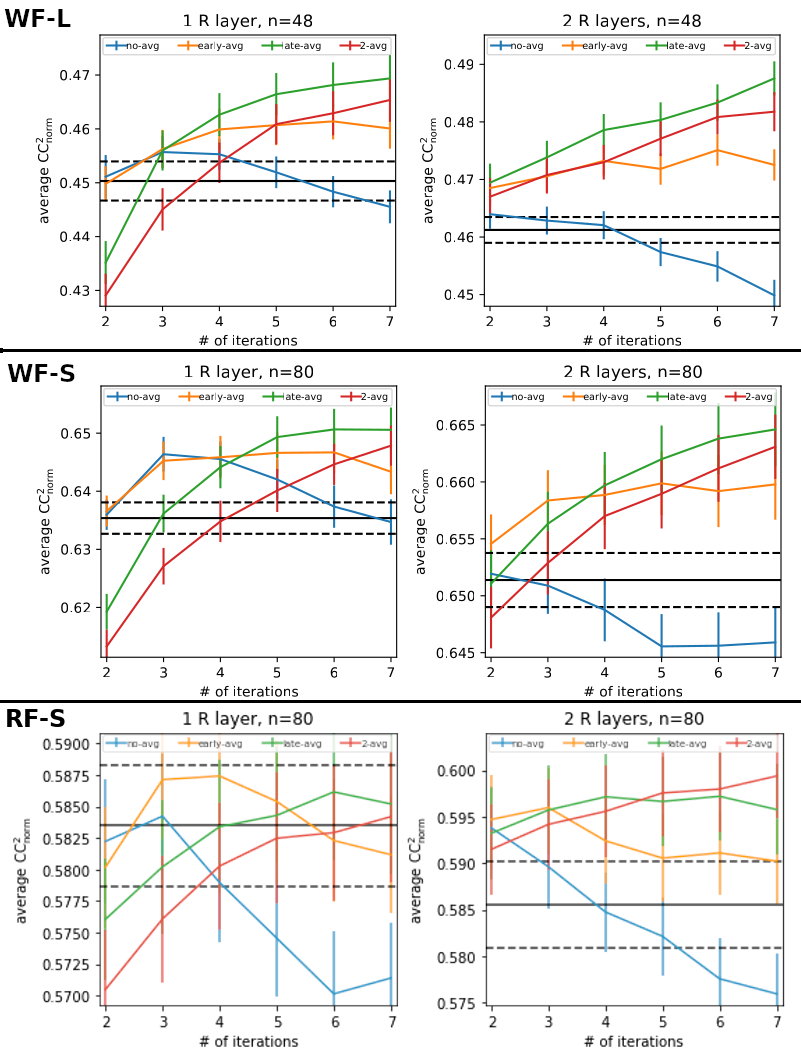}
    \caption{Average performance on each dataset as a function of number of iterations, for models divided by their number of recurrent layers.}
    \label{perf_length_layers_split}
\end{figure}

Figure~\ref{perf_length_layers_split} shows performance of recurrent models as a function of number of iterations, with parameter-matched feedforward models' performance shown as a horizontal line. It is analogous to Figure 3 of the main paper, but divides the results up between models with one and two recurrent layers. Two-recurrent-layer models tend to further outperform their parameter-matched feedforward models, particularly for the {\bf RF-S} data, where one-recurrent-layer models are no better than feedforward.

\section{Path length distributions for one-recurrent-layer models}

\begin{figure}[htb]
    \centering
    \includegraphics[width=0.7\textwidth]{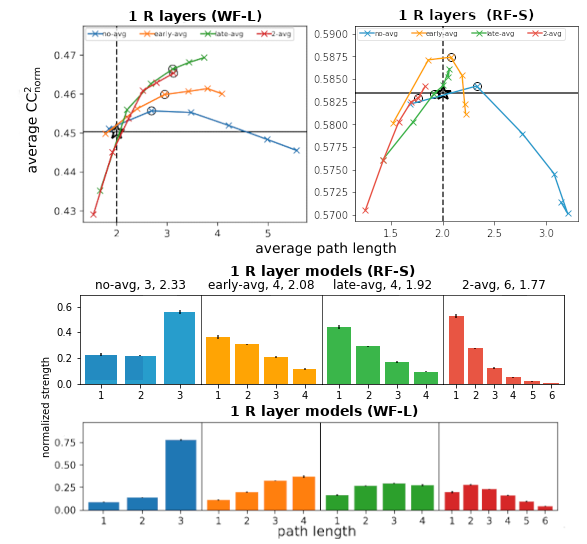}
    \caption{Performance plotted against average path length, and path length distributions for one-layer recurrent models trained on the {\bf RF-S} and {\bf WF-L} datasets. For these models as well as the two-recurrent-layer ones, training on the {\bf RF-S} dataset results in higher weighting of shorter paths.}
    \label{fig5_1r_model_supplementary}
\end{figure}

Figure~\ref{fig5_1r_model_supplementary} shows the performance versus path length and path length distributions for models with one recurrent layer, analogous to the results for two-recurrent layer models shown in Figure 5 of the main paper. Ablation studies were not performed on these models.

\section{Path length distributions for {\bf WF-S} models}
\label{nips2021:appendix:8k_distributions}

\begin{figure}[htb]
    \centering
    \includegraphics[width=0.7\textwidth]{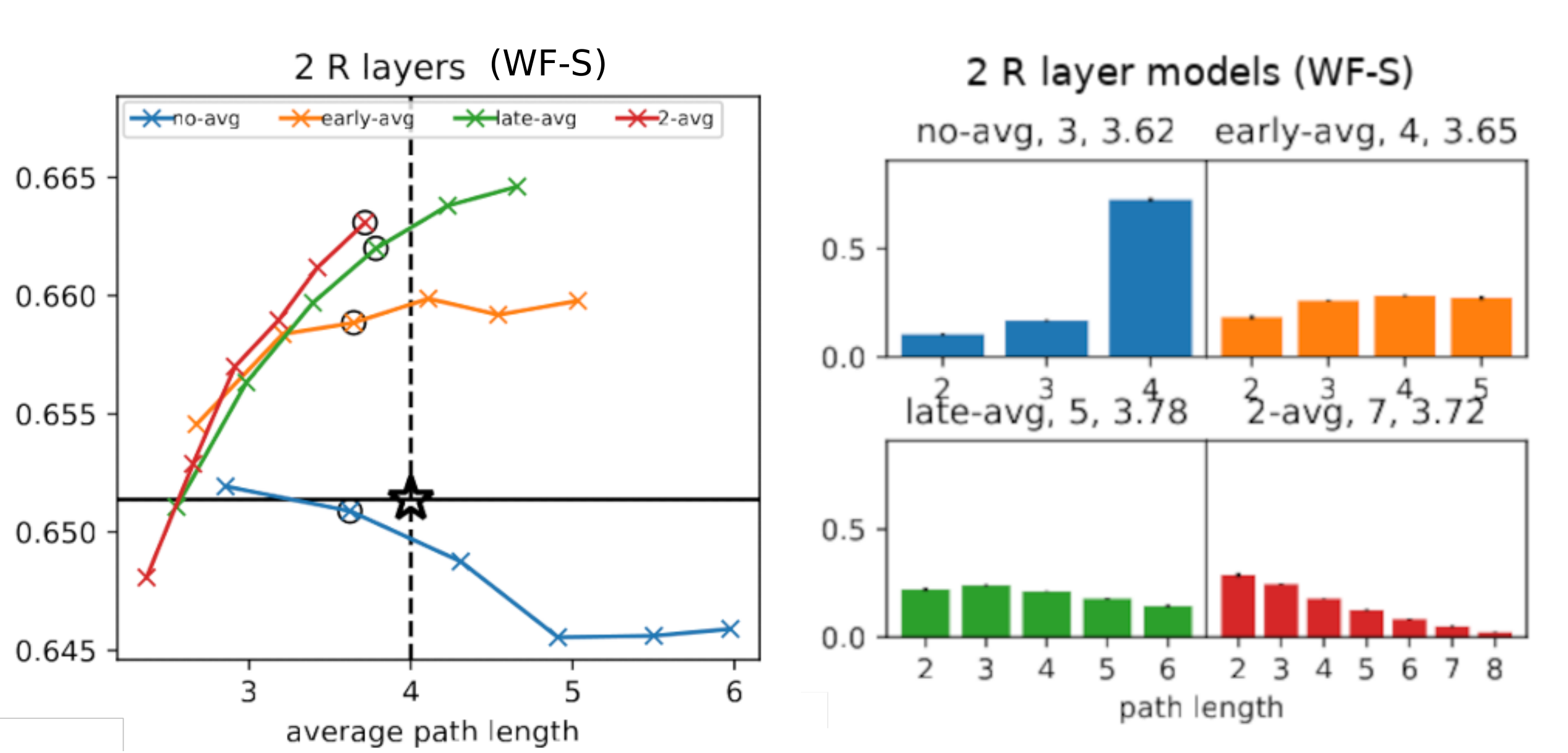}
    \caption{Path length and diversity for models trained on the {\bf WF-S} data. Analogous to panels {\bf A, B, D, E} of Figure 5 in the main paper, but for the other dataset.}
    \label{8k_path_length_distributions}
\end{figure}

Figure~\ref{8k_path_length_distributions} shows the performance as a function of path length and path length distributions for each 2-layer model readout mode trained on the {\bf WF-S} data. Notably, the path length distributions fall somewhere in between those observed in the {\bf RF-S} and {\bf WF-L} datasets in the main paper's figure, with shorter paths having a higher impact than for {\bf WF-L} models, but a flatter distribution than the {\bf RF-S} models. The {\bf WF-S} data has wide-field images that allow for surround modulation (like the {\bf WF-L} data), but short presentations that limit the amount of modulation possible (like the {\bf RF-S}).

\section{Multi-path analysis for V1-only WF-L/WF-S models}
\label{nips2021:appendix:multipath_v1_only}

\begin{figure}[htb]
    \centering
    \includegraphics[width=0.99\textwidth]{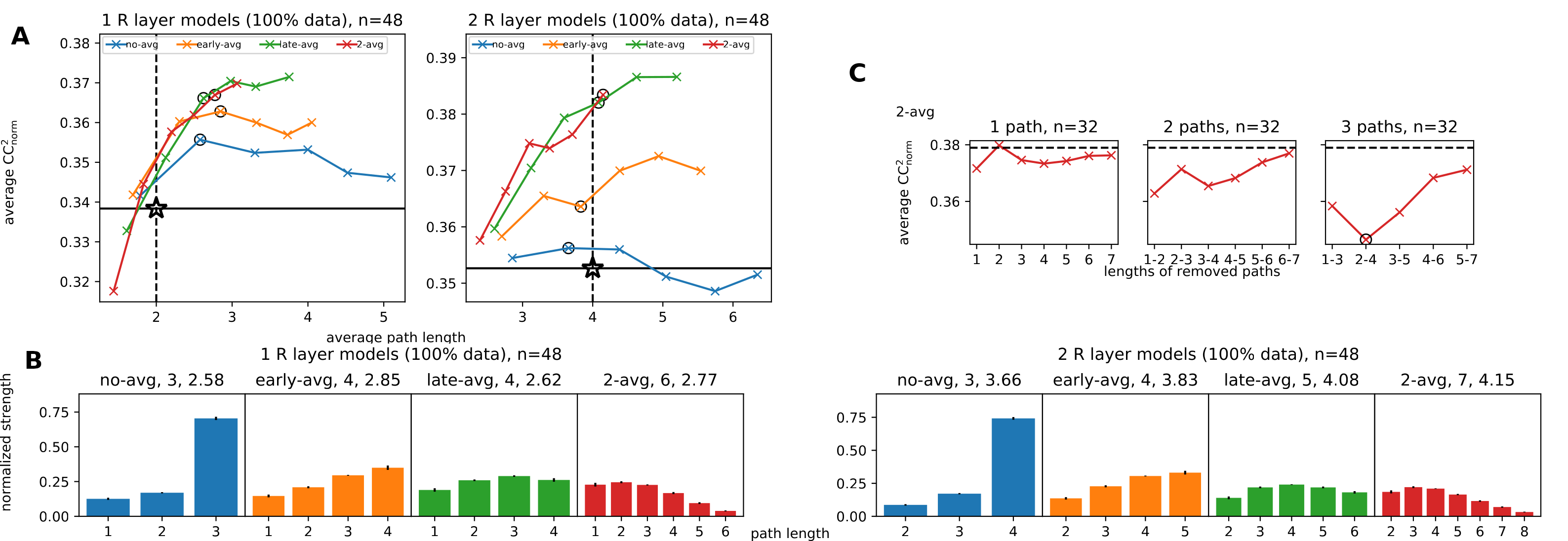}
    \caption{Performance versus average path length (A), path length distributions (B), and ablation results (C) for models trained on only the V1 neurons in the {\bf WF-L} dataset. As in Figure 5 and elsewhere, ablation studies were only performed on the 2-recurrent-layer models.}
    \label{fig5_gayav1_supplementary}
\end{figure}

The {\bf WF-L} dataset (as well as the {\bf WF-S} dataset) consist of both V1 and V2 neurons, while the {\bf RF-S} dataset contains only V1 neurons. To see whether the differences between models trained on the {\bf WF-L} and {\bf RF-S} in Figure 5 of the main paper depend more on the stimulation paradigm or the cortical area of the modeled neurons, we trained models on a subset of the {\bf WF-L} dataset containing only neurons in V1, 6 out of the 34 neurons. Figure~\ref{fig5_gayav1_supplementary} shows the analyses of Figure 5 for both 1-recurrent-layer and 2-recurrent-layer models trained on this V1-only {\bf WF-L} subset. The results generally resemble the results for the full {\bf WF-L} dataset in Figure 5 of the main paper (2-recurrent-layer models) and Figure~\ref{fig5_1r_model_supplementary} (1-layer recurrent models), with path length distributions notably preferring longer paths than those for models trained on the {\bf RF-S} data in Figure 5. However, the ablation study results are in-between, with the removal of mid-length paths having the largest effect when 3 paths are removed (like the full {\bf WF-L}), but the shortest paths mattering most when only 1 or 2 paths are removed (more like the {\bf RF-S}). This suggests that the differences we see between models trained on the {\bf WF-L} and {\bf RF-S} are substantially due to the difference in presentation paradigms, but in part driven by the different cortical areas modeled.

\begin{figure}[htb]
    \centering
    \includegraphics[width=0.55\textwidth]{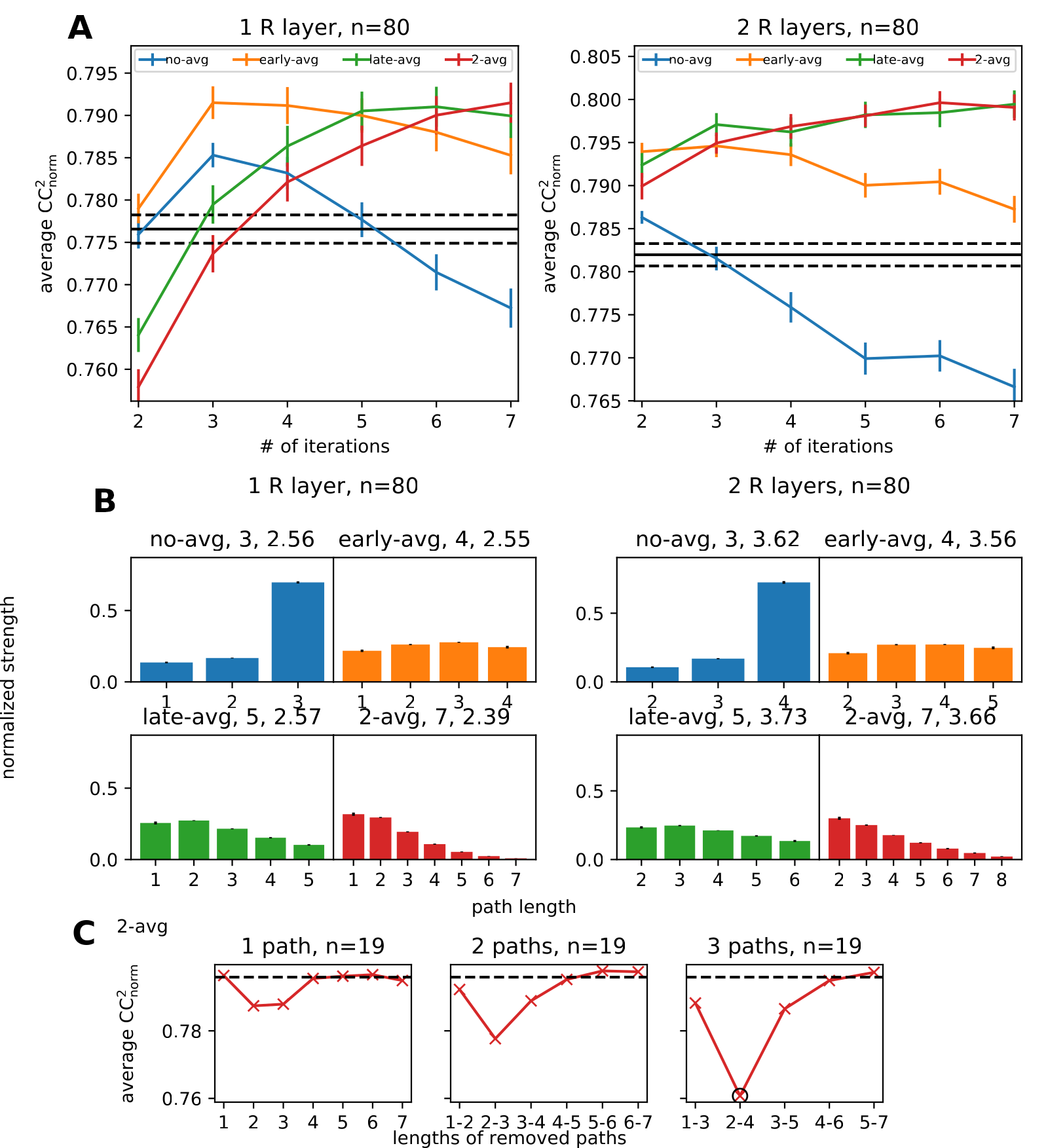}
    \caption{Performance versus average path length (A), path length distributions (B), and ablation results (C) for models trained on only the V1 neurons in the {\bf WF-S} dataset. As in Figure 5 and elsewhere, ablation studies were only performed on the 2-recurrent-layer models. Comparable to Figure~\ref{8k_path_length_distributions}, but only training on V1 neurons.}
    \label{fig5_8kv1_supplementary}
\end{figure}

In addition, while we did not include {\bf WF-S}-trained models in Figure 5 of the main paper, we replicated those results (included in Figure~\ref{8k_path_length_distributions}) on models trained on the subset of the {\bf WF-S} data containing only V1 cells. The results, displayed in Figure~\ref{fig5_8kv1_supplementary}, are fairly similar to those of models trained on all the {\bf WF-S} neurons, including the V2 cells. This supports our claims that the difference we see in the importance of paths of different lengths between datasets is driven mostly, but not entirely, by the presentation paradigm of the stimuli than by the cortical area modeled.

\section{Ablation study}
\label{appendix:ablation}

\begin{figure}[htb]
    \centering
    \includegraphics[width=0.75\textwidth]{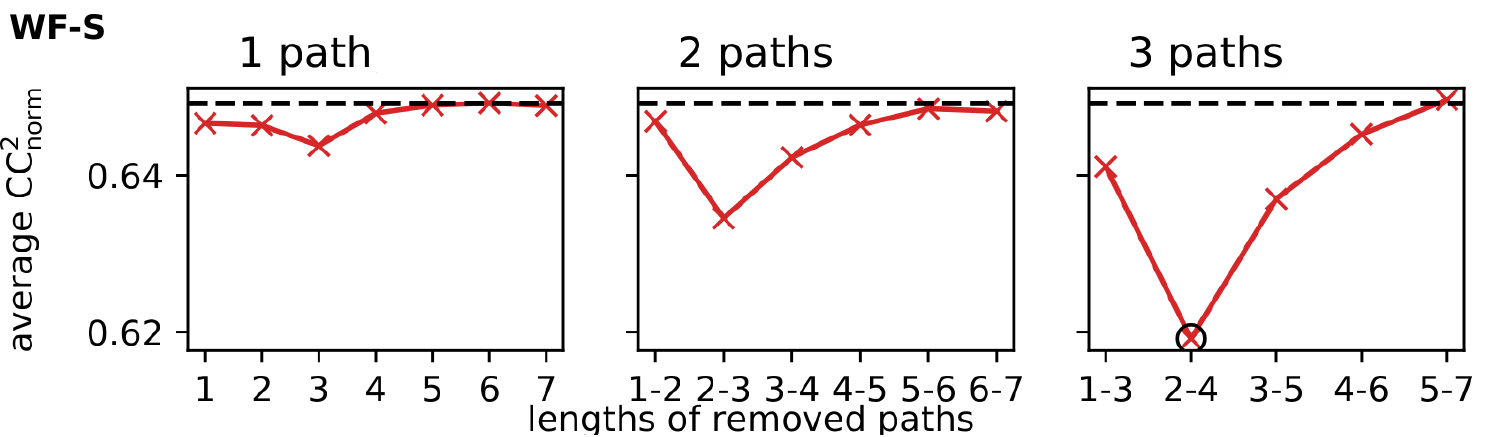}
    \caption{Performance metrics of ablated 2-layer, 7-iteration models of 16 and 32 channels on the {\bf WF-S} dataset, with 1, 2, or 3 paths removed at a time. Each panel shows results of \texttt{2-avg} models with 1, 2, 3 paths removed from left to right. Dotted lines show the performance metrics of full models that have all paths. Each circle denotes the worst performing ablation configuration with 3 paths removed.}
    \label{fig:results:8k:ablation7}
\end{figure}

To compute the baseline shown as horizontal dashed lines in the main paper's Figure 5{\bf C} and~\ref{fig:results:8k:ablation7}, we trained the multi-path models corresponding to the original recurrent models with 2 layers, 7 iterations, and 16 or 32 channels using 100\% of training data. 
We used all the training data because the reformulation of recurrent models as multi-path models worked better with more training data, and we did not train recurrent models with other numbers of layers or channels because models with 3 layers or more than 32 channels exceeded GPU memory limits during training and 8-channel recurrent models did not outperform feed-forward models much. We also restricted this analysis to the \texttt{2-avg} models, since the neurophysiological experiments and performance values highlighted those as the most relevant.


To compute the performance metrics of ablated models in the figures, we trained additional multi-path models like those for getting the baseline, but with paths of certain lengths removed. In each of the figures, the left panel shows the performance metric change when paths of a particular length (1,2,3,4,5,6) were removed; the middle panel shows the performance metrics change when paths of two adjacent lengths (1-2, 2-3, 3-4, 4-5, 5-6) were removed; the right panel shows the performance metric change when paths of three adjacent lengths (1-3, 2-4, 3-5, 4-6, 5-7) were removed. 

\FloatBarrier

\section{Extensive comparison of  recurrent models against  feed-forward models}
\label{nips2021:results_additional_ff_vs_r}
\label{appendix:ff_model_vs_r}

The scatterplots in this section show the performance of 24 types of recurrent models against hyperparameter-matched feedforward models for the three datasets, split up by readout mode and number of iterations. Each point is the average performance value of two random initializations for a model with a particular set of hyperparameters.  
For the {\bf RF-S} dataset, the scatter plots include both 1R-layer and 2R-layer models, like they do for the other datasets. Upon scrutiny, we found 1R recurrent models do not yield a statistically significant improvement on this dataset, but 2R recurrent models do, as shown in the histograms of Figure~\ref{histogram_fig}, where much of the mass is concentrated above zero.
Results are also shown for models trained subsets of the {\bf WF-S} and {\bf WF-L} datasets containing only V1 neurons, for models trained on 100\% of the data only. They continue to show the improvement of recurrent networks.


\begin{figure}[htb]
    \centering
    \includegraphics[width=0.8\textwidth]{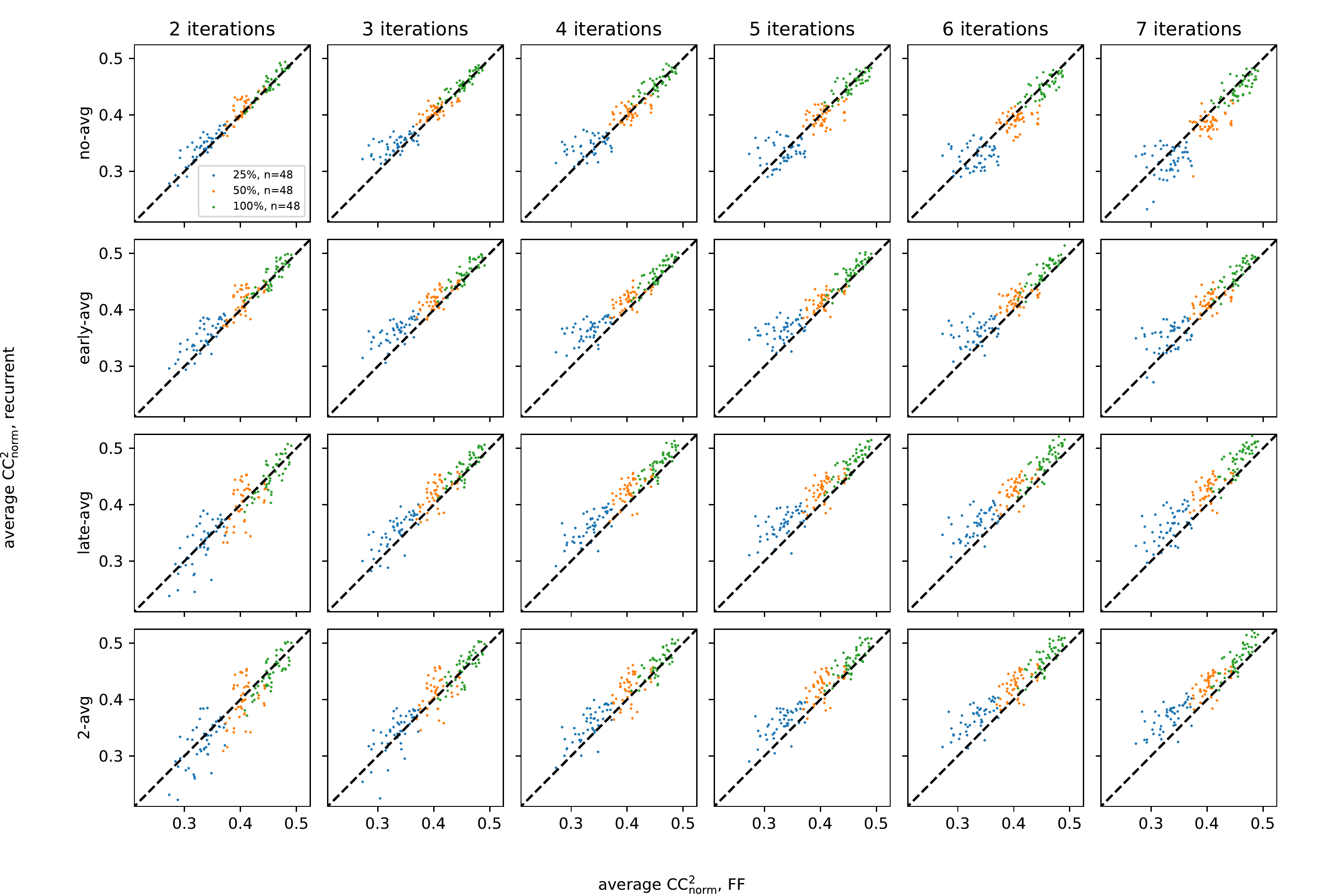}
    \caption{
    Prediction Performance for the {\bf WF-L} trained models:  Twenty four varieties (24 combinations of readout modes and number of iterations) of recurrent models vs.\@ matched-sized feed-forward models under \num{24}  combinations of readout mode and number of iterations, for the {\bf WF-L}  data. Panels in the same row have the same readout mode as indicated on the left of the whole figure, and panels in the same column have the same number of iterations as indicated on the top of the whole figure. While we trained \num{96} feed-forward models and their corresponding recurrent models under \num{24} conditions  for each explored training data size, each panel only shows $n=48$ pairs of (aggregated) recurrent vs.\@ feed-forward models per training data size, as we averaged the results over the two explored model parameter initialization seeds for more robust metrics 
    In each panel, model pairs trained under different amounts of training data (25\%, 50\%, 100\%) are shown in different colors, with \num{48} (aggregated) model pairs for each explored  training data size.}
    \label{fig:results:tang:scatter_r_vs_ff:3rd}
\end{figure}



\begin{figure}[htb]
    \centering
    \includegraphics[width=0.8\textwidth]{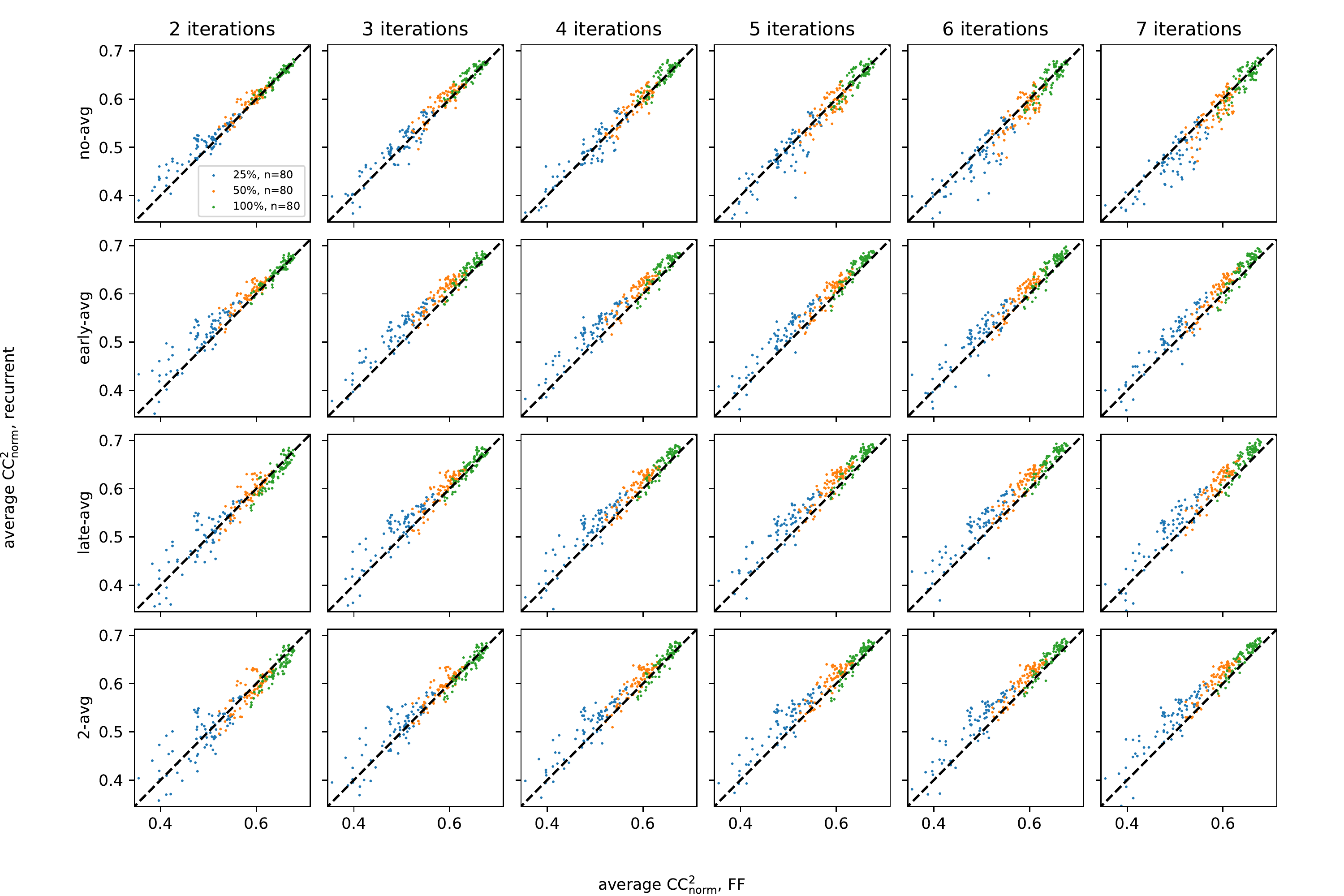}
    \caption{Predictive performance for the {\bf WF-S} trained models: Same as Figure~\ref{fig:results:tang:scatter_r_vs_ff:3rd} but for the {\bf WF-S} data.}
    \label{fig:results:8k:scatter_r_vs_ff:3rd}
\end{figure}

\begin{figure}[htb]
    \centering
    \includegraphics[width=0.8\textwidth]{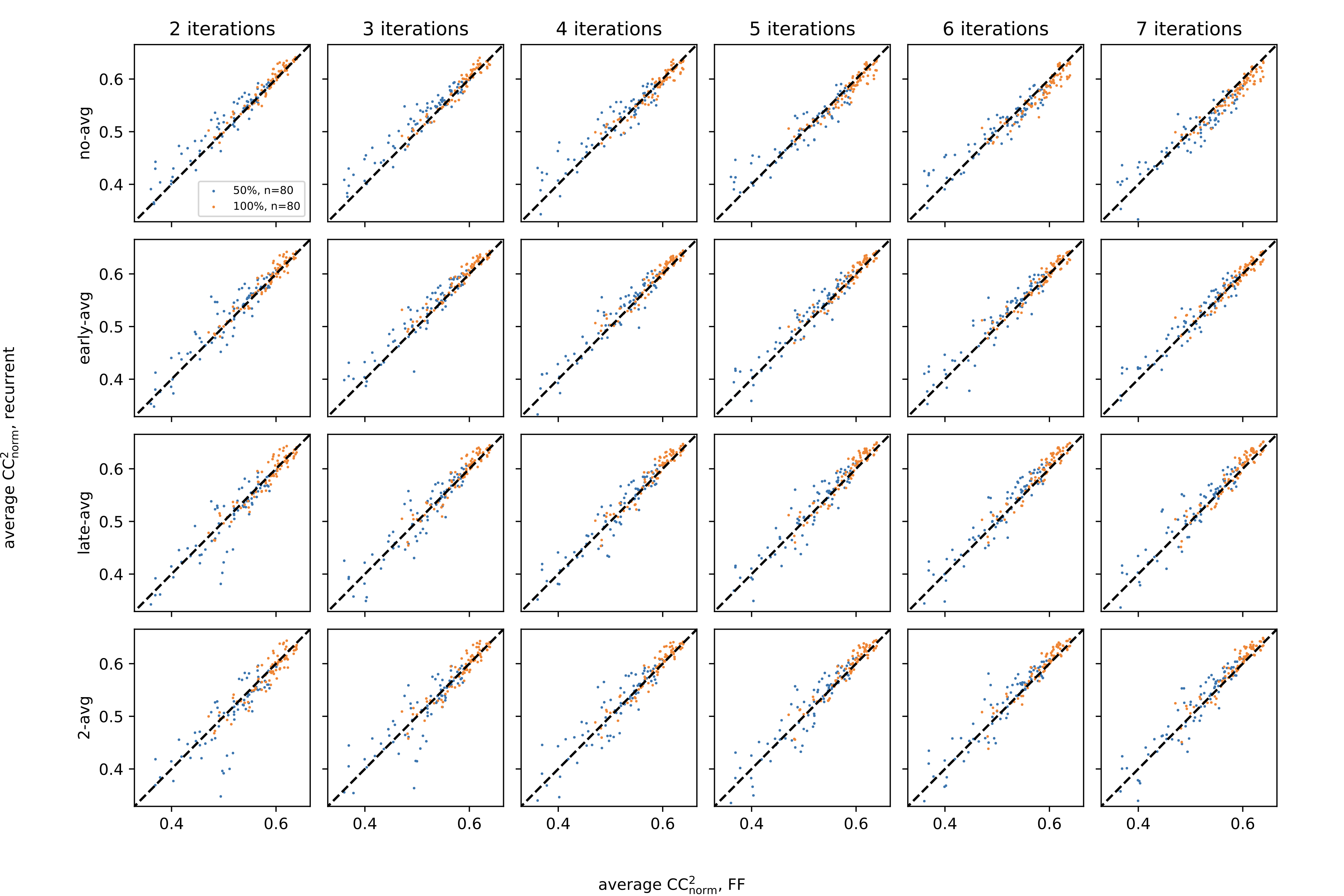}
    \caption{Predictive performance for the {\bf RF-S} trained models: Same as Figure~\ref{fig:results:tang:scatter_r_vs_ff:3rd} but for {\bf RF-S} data. Note only the 50\% and 100\% cases are shown as the 25\% data are too impoverished, 1000 images with 3-4 repeats are too noisy, resulting in performance metric below 0.3 for all models, too poor for meaningful comparison. }
\end{figure}


\begin{figure}[htb]
    \centering
    \includegraphics[width=0.8\textwidth]{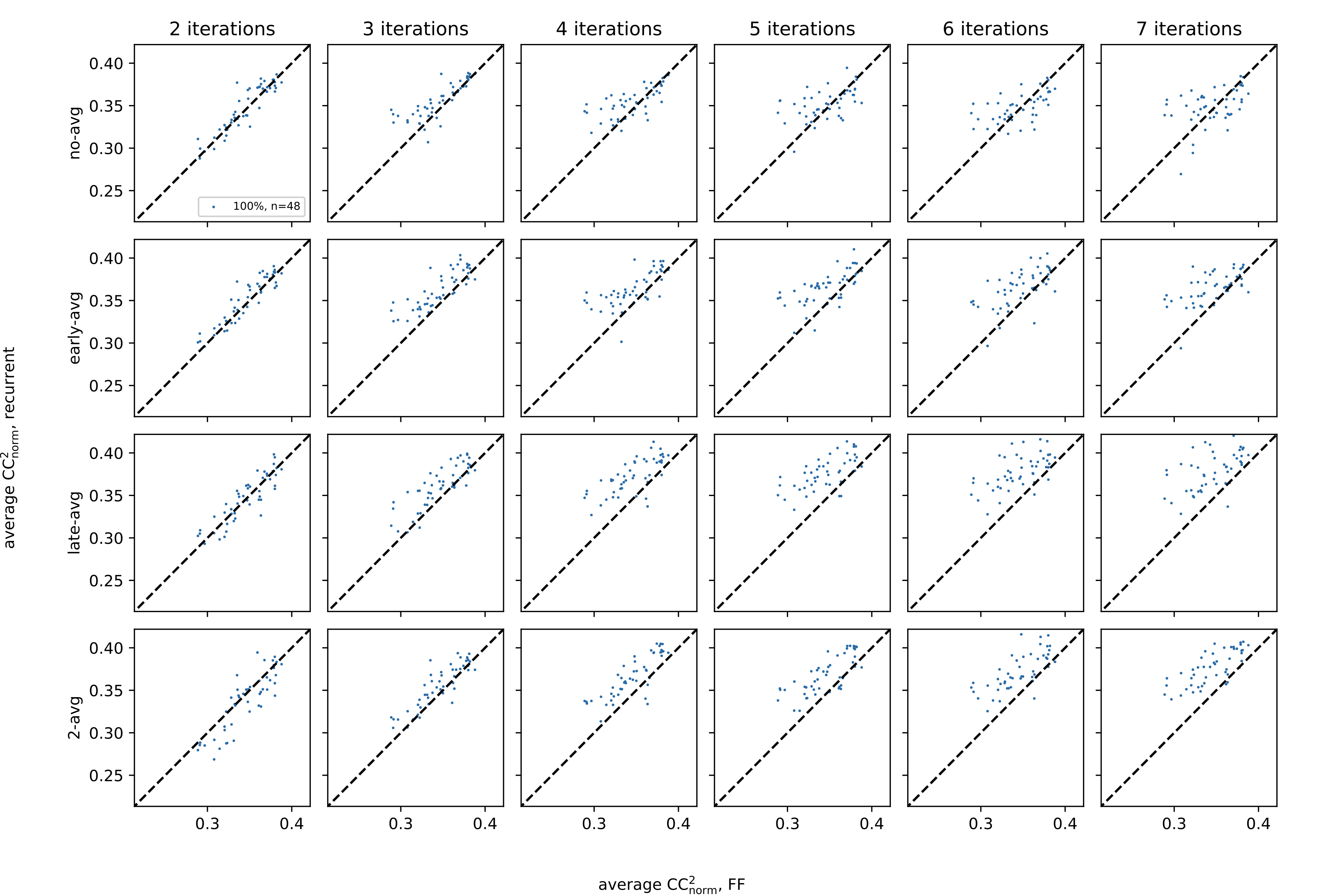}
    \caption{Predictive performance for the {\bf WF-L}-V1-only-trained models, trained with 100\% data.}
\end{figure}

\begin{figure}[htb]
    \centering
    \includegraphics[width=0.8\textwidth]{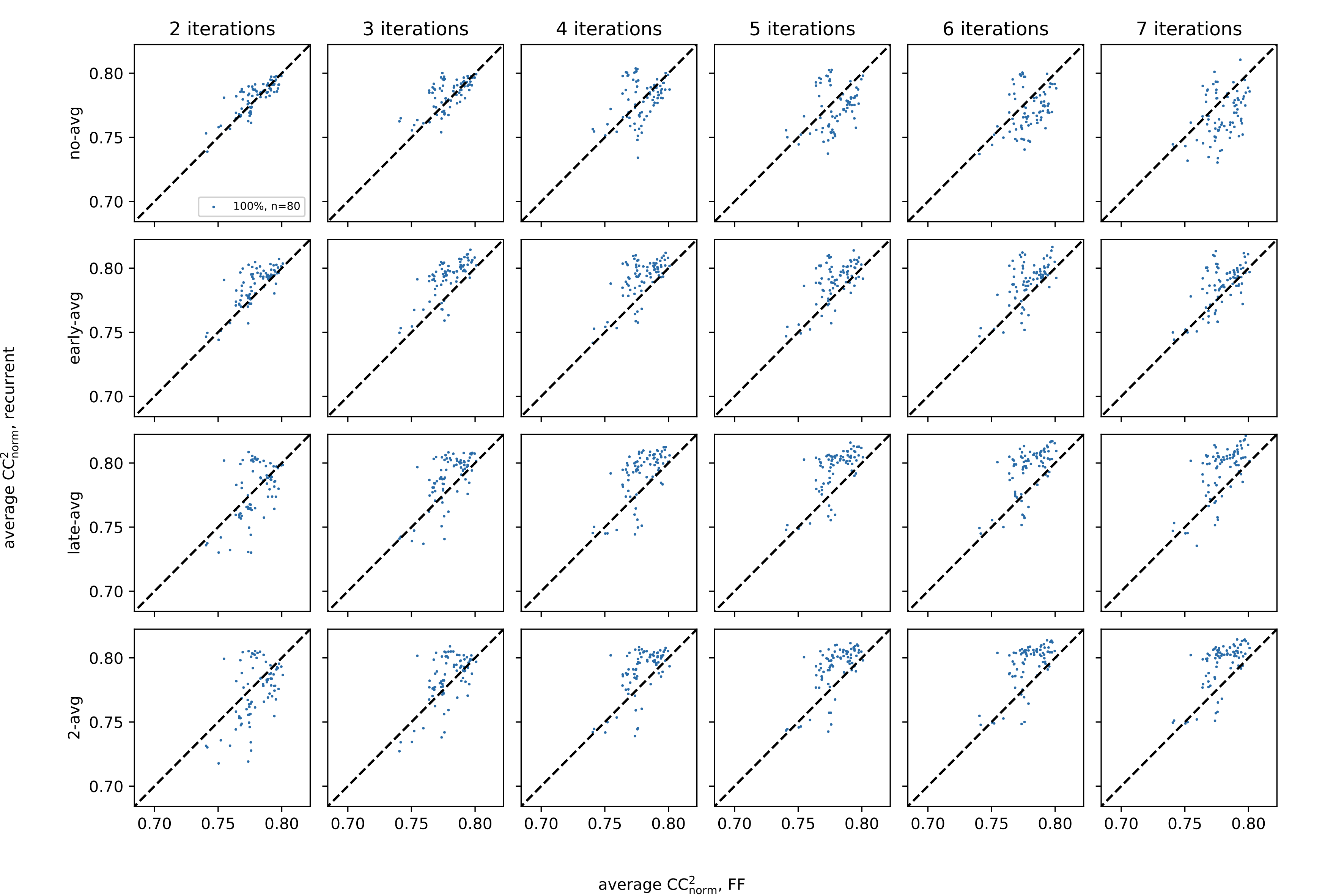}
    \caption{Predictive performance for the {\bf WF-S}-V1-only-trained models, trained with 100\% data.}
\end{figure}

\begin{figure}[htb]
    \centering
    \includegraphics[width=0.8\textwidth]{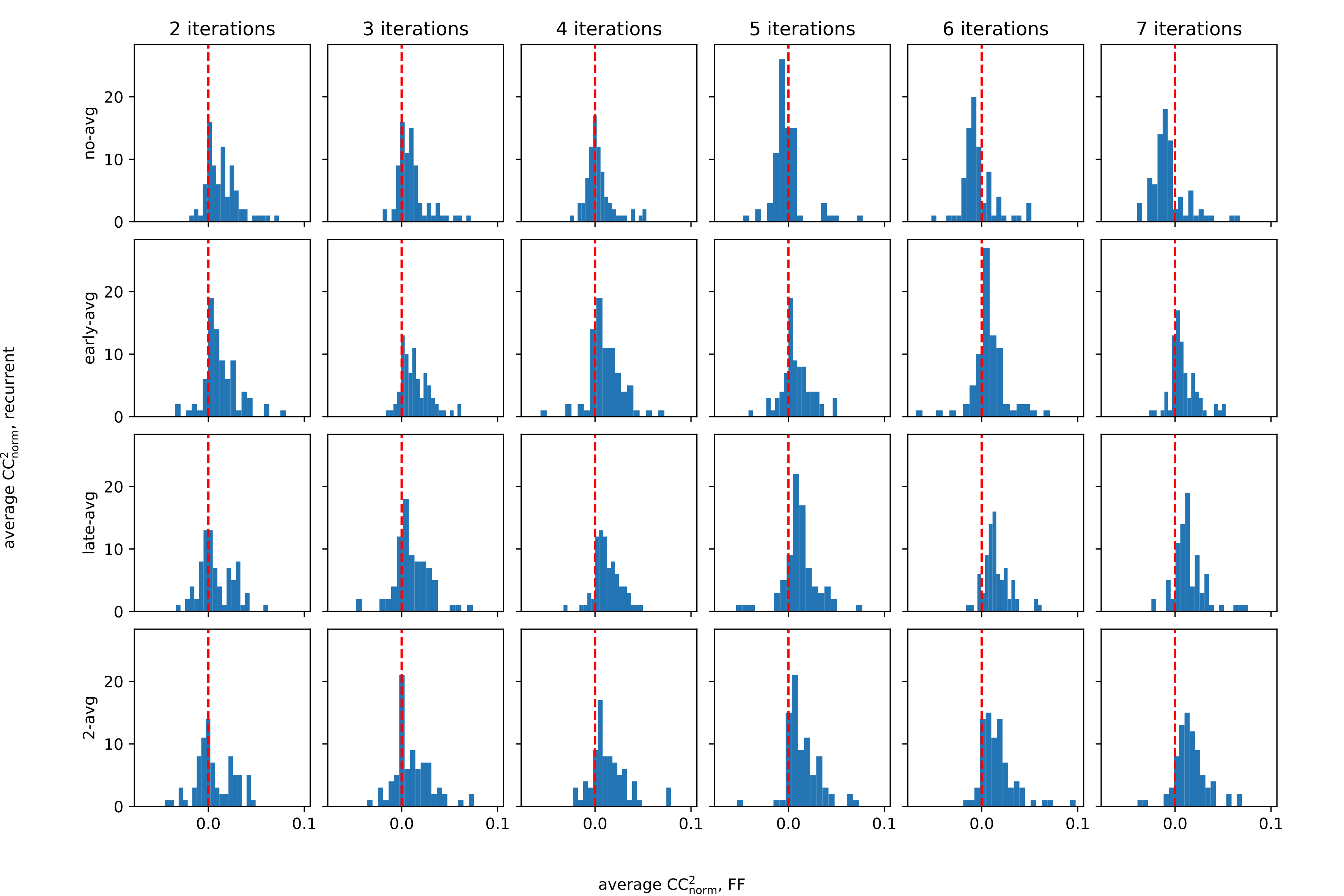}
    \caption{Performance improvement of 2-recurrent-layer recurrent models against feedforward models trained for the {\bf RF-S} dataset. Histograms of (R-F)/F performance metric for the 24 varieties of recurrent models. R is the average $\ccnorm$ for the recurrent models and F is the average $\ccnorm$ for the matched-size feedforward model. }
    \label{histogram_fig}
\end{figure}


%% file: main.bbl
\begin{thebibliography}{35}
\providecommand{\natexlab}[1]{#1}
\providecommand{\url}[1]{\texttt{#1}}
\expandafter\ifx\csname urlstyle\endcsname\relax
  \providecommand{\doi}[1]{doi: #1}\else
  \providecommand{\doi}{doi: \begingroup \urlstyle{rm}\Url}\fi

\bibitem[Kindel et~al.(2019)Kindel, Christensen, and
  Zylberberg]{10.1167/19.4.29}
William~F. Kindel, Elijah~D. Christensen, and Joel Zylberberg.
\newblock {Using deep learning to probe the neural code for images in primary
  visual cortex}.
\newblock \emph{Journal of Vision}, 19\penalty0 (4):\penalty0 29--29, 04 2019.
\newblock ISSN 1534-7362.
\newblock \doi{10.1167/19.4.29}.
\newblock URL \url{https://doi.org/10.1167/19.4.29}.

\bibitem[Klindt et~al.(2017)Klindt, Ecker, Euler, and
  Bethge]{DBLP:conf/nips/KlindtEEB17}
David Klindt, Alexander~S. Ecker, Thomas Euler, and Matthias Bethge.
\newblock Neural system identification for large populations separating "what"
  and "where".
\newblock In Isabelle Guyon, Ulrike von Luxburg, Samy Bengio, Hanna~M. Wallach,
  Rob Fergus, S.~V.~N. Vishwanathan, and Roman Garnett, editors, \emph{Advances
  in Neural Information Processing Systems 30: Annual Conference on Neural
  Information Processing Systems 2017, 4-9 December 2017, Long Beach, CA,
  {USA}}, pages 3509--3519, 2017.
\newblock URL
  \url{http://papers.nips.cc/paper/6942-neural-system-identification-for-large-populations-separating-what-and-where}.

\bibitem[Yamins and DiCarlo(2016)]{Yamins:2016hg}
Daniel L~K Yamins and James~J DiCarlo.
\newblock {Using goal-driven deep learning models to understand sensory
  cortex}.
\newblock \emph{Nature Neuroscience}, 19\penalty0 (3):\penalty0 356--365,
  February 2016.
\newblock \doi{10.1038/nn.4244}.
\newblock URL \url{http://www.nature.com/doifinder/10.1038/nn.4244}.

\bibitem[Zhang et~al.(2019)Zhang, Lee, Li, Liu, and Tang]{Zhang2019}
Yimeng Zhang, Tai~Sing Lee, Ming Li, Fang Liu, and Shiming Tang.
\newblock Convolutional neural network models of v1 responses to complex
  patterns.
\newblock \emph{Journal of Computational Neuroscience}, 46\penalty0
  (1):\penalty0 33--54, Feb 2019.
\newblock ISSN 1573-6873.
\newblock \doi{10.1007/s10827-018-0687-7}.
\newblock URL \url{https://doi.org/10.1007/s10827-018-0687-7}.

\bibitem[Cadena et~al.(2019)Cadena, Denfield, Walker, Gatys, Tolias, Bethge,
  and Ecker]{10.1371/journal.pcbi.1006897}
Santiago~A. Cadena, George~H. Denfield, Edgar~Y. Walker, Leon~A. Gatys,
  Andreas~S. Tolias, Matthias Bethge, and Alexander~S. Ecker.
\newblock Deep convolutional models improve predictions of macaque v1 responses
  to natural images.
\newblock \emph{PLOS Computational Biology}, 15\penalty0 (4):\penalty0 1--27,
  04 2019.
\newblock \doi{10.1371/journal.pcbi.1006897}.
\newblock URL \url{https://doi.org/10.1371/journal.pcbi.1006897}.

\bibitem[Kriegeskorte(2015)]{doi:10.1146/annurev-vision-082114-035447}
Nikolaus Kriegeskorte.
\newblock Deep neural networks: A new framework for modeling biological vision
  and brain information processing.
\newblock \emph{Annual Review of Vision Science}, 1\penalty0 (1):\penalty0
  417--446, 2015.
\newblock \doi{10.1146/annurev-vision-082114-035447}.
\newblock URL \url{https://doi.org/10.1146/annurev-vision-082114-035447}.
\newblock PMID: 28532370.

\bibitem[Felleman and Van~Essen(1991)]{10.1093/cercor/1.1.1-a}
Daniel~J. Felleman and David~C. Van~Essen.
\newblock {Distributed Hierarchical Processing in the Primate Cerebral Cortex}.
\newblock \emph{Cerebral Cortex}, 1\penalty0 (1):\penalty0 1--47, 01 1991.
\newblock ISSN 1047-3211.
\newblock \doi{10.1093/cercor/1.1.1-a}.
\newblock URL \url{https://doi.org/10.1093/cercor/1.1.1-a}.

\bibitem[Markov et~al.(2014)Markov, Vezoli, Chameau, Falchier, Quilodran,
  Huissoud, Lamy, Misery, Giroud, Ullman, Barone, Dehay, Knoblauch, and
  Kennedy]{10.1002/cne.23458}
Nikola~T. Markov, Julien Vezoli, Pascal Chameau, Arnaud Falchier, Ren{\'e}
  Quilodran, Cyril Huissoud, Camille Lamy, Pierre Misery, Pascale Giroud,
  Shimon Ullman, Pascal Barone, Colette Dehay, Kenneth Knoblauch, and Henry
  Kennedy.
\newblock Anatomy of hierarchy: Feedforward and feedback pathways in macaque
  visual cortex.
\newblock \emph{Journal of Comparative Neurology}, 522\penalty0 (1):\penalty0
  225--259, 2014.
\newblock \doi{https://doi.org/10.1002/cne.23458}.
\newblock URL \url{https://onlinelibrary.wiley.com/doi/abs/10.1002/cne.23458}.

\bibitem[Han et~al.(2018)Han, Wen, Zhang, Fu, Culurciello, and
  Liu]{DBLP:conf/nips/HanWZFCL18}
Kuan Han, Haiguang Wen, Yizhen Zhang, Di~Fu, Eugenio Culurciello, and Zhongming
  Liu.
\newblock Deep predictive coding network with local recurrent processing for
  object recognition.
\newblock In Samy Bengio, Hanna~M. Wallach, Hugo Larochelle, Kristen Grauman,
  Nicol{\`{o}} Cesa{-}Bianchi, and Roman Garnett, editors, \emph{Advances in
  Neural Information Processing Systems 31: Annual Conference on Neural
  Information Processing Systems 2018, NeurIPS 2018, 3-8 December 2018,
  Montr{\'{e}}al, Canada.}, pages 9221--9233, 2018.
\newblock URL
  \url{http://papers.nips.cc/paper/8133-deep-predictive-coding-network-with-local-recurrent-processing-for-object-recognition}.

\bibitem[Nayebi et~al.(2018)Nayebi, Bear, Kubilius, Kar, Ganguli, Sussillo,
  DiCarlo, and Yamins]{nayebi2018task}
Aran Nayebi, Daniel Bear, Jonas Kubilius, Kohitij Kar, Surya Ganguli, David
  Sussillo, James~J. DiCarlo, and Daniel L.~K. Yamins.
\newblock Task-driven convolutional recurrent models of the visual system.
\newblock \emph{CoRR}, abs/1807.00053, 2018.
\newblock URL \url{http://arxiv.org/abs/1807.00053}.

\bibitem[Kubilius et~al.(2019)Kubilius, Schrimpf, Hong, Majaj, Rajalingham,
  Issa, Kar, Bashivan, Prescott{-}Roy, Schmidt, Nayebi, Bear, Yamins, and
  DiCarlo]{DBLP:conf/nips/KubiliusSHMRIKB19}
Jonas Kubilius, Martin Schrimpf, Ha~Hong, Najib~J. Majaj, Rishi Rajalingham,
  Elias~B. Issa, Kohitij Kar, Pouya Bashivan, Jonathan Prescott{-}Roy, Kailyn
  Schmidt, Aran Nayebi, Daniel Bear, Daniel~L. Yamins, and James~J. DiCarlo.
\newblock Brain-like object recognition with high-performing shallow recurrent
  anns.
\newblock In Hanna~M. Wallach, Hugo Larochelle, Alina Beygelzimer, Florence
  d'Alch{\'{e}}{-}Buc, Emily~B. Fox, and Roman Garnett, editors, \emph{Advances
  in Neural Information Processing Systems 32: Annual Conference on Neural
  Information Processing Systems 2019, NeurIPS 2019, December 8-14, 2019,
  Vancouver, BC, Canada}, pages 12785--12796, 2019.
\newblock URL
  \url{https://proceedings.neurips.cc/paper/2019/hash/7813d1590d28a7dd372ad54b5d29d033-Abstract.html}.

\bibitem[Opitz and Maclin(1999)]{opitz1999}
David Opitz and Richard Maclin.
\newblock Popular ensemble methods: An empirical study.
\newblock \emph{J. Artif. Int. Res.}, 11\penalty0 (1):\penalty0 169‚Äì198,
  jul 1999.
\newblock ISSN 1076-9757.

\bibitem[Polikar(2006)]{Polikar2006}
R.~Polikar.
\newblock Ensemble based systems in decision making.
\newblock \emph{IEEE Circuits and Systems Magazine}, 6\penalty0 (3):\penalty0
  21--45, 2006.
\newblock \doi{10.1109/MCAS.2006.1688199}.

\bibitem[Rokach(2010)]{rokach2010}
Lior Rokach.
\newblock Ensemble-based classifiers.
\newblock \emph{Artificial Intelligence Review}, 33\penalty0 (1):\penalty0
  1--39, 2010.
\newblock \doi{10.1007/s10462-009-9124-7}.
\newblock URL \url{https://doi.org/10.1007/s10462-009-9124-7}.

\bibitem[Spoerer et~al.(2020)Spoerer, Kietzmann, Mehrer, Charest, and
  Kriegeskorte]{10.1371/journal.pcbi.1008215}
Courtney~J. Spoerer, Tim~C. Kietzmann, Johannes Mehrer, Ian Charest, and
  Nikolaus Kriegeskorte.
\newblock Recurrent neural networks can explain flexible trading of speed and
  accuracy in biological vision.
\newblock \emph{PLOS Computational Biology}, 16\penalty0 (10):\penalty0 1--27,
  10 2020.
\newblock \doi{10.1371/journal.pcbi.1008215}.
\newblock URL \url{https://doi.org/10.1371/journal.pcbi.1008215}.

\bibitem[Spoerer et~al.(2017)Spoerer, McClure, and
  Kriegeskorte]{10.3389/fpsyg.2017.01551}
Courtney~J. Spoerer, Patrick McClure, and Nikolaus Kriegeskorte.
\newblock Recurrent convolutional neural networks: A better model of biological
  object recognition.
\newblock \emph{Frontiers in Psychology}, 8:\penalty0 1551, 2017.
\newblock ISSN 1664-1078.
\newblock \doi{10.3389/fpsyg.2017.01551}.
\newblock URL
  \url{https://www.frontiersin.org/article/10.3389/fpsyg.2017.01551}.

\bibitem[McIntosh et~al.(2017)McIntosh, Maheswaranathan, Nayebi, Ganguli, and
  Baccus]{McIntosh:2017uc}
Lane~T McIntosh, Niru Maheswaranathan, Aran Nayebi, Surya Ganguli, and
  Stephen~A Baccus.
\newblock {Deep Learning Models of the Retinal Response to Natural Scenes}.
\newblock \emph{ArXiv e-prints}, q-bio.NC, February 2017.
\newblock URL \url{https://arxiv.org/abs/1702.01825}.

\bibitem[Bai et~al.(2019)Bai, Kolter, and Koltun]{DBLP:conf/nips/BaiKK19}
Shaojie Bai, J.~Zico Kolter, and Vladlen Koltun.
\newblock Deep equilibrium models.
\newblock In Hanna~M. Wallach, Hugo Larochelle, Alina Beygelzimer, Florence
  d'Alch{\'{e}}{-}Buc, Emily~B. Fox, and Roman Garnett, editors, \emph{Advances
  in Neural Information Processing Systems 32: Annual Conference on Neural
  Information Processing Systems 2019, NeurIPS 2019, December 8-14, 2019,
  Vancouver, BC, Canada}, pages 688--699, 2019.
\newblock URL
  \url{https://proceedings.neurips.cc/paper/2019/hash/01386bd6d8e091c2ab4c7c7de644d37b-Abstract.html}.

\bibitem[Huang et~al.(2017)Huang, Liu, van~der Maaten, and
  Weinberger]{DBLP:conf/cvpr/HuangLMW17}
Gao Huang, Zhuang Liu, Laurens van~der Maaten, and Kilian~Q. Weinberger.
\newblock Densely connected convolutional networks.
\newblock In \emph{2017 {IEEE} Conference on Computer Vision and Pattern
  Recognition, {CVPR} 2017, Honolulu, HI, USA, July 21-26, 2017}, pages
  2261--2269. {IEEE} Computer Society, 2017.
\newblock \doi{10.1109/CVPR.2017.243}.
\newblock URL \url{https://doi.org/10.1109/CVPR.2017.243}.

\bibitem[Veit et~al.(2016)Veit, Wilber, and Belongie]{NIPS2016_37bc2f75}
Andreas Veit, Michael~J Wilber, and Serge Belongie.
\newblock Residual networks behave like ensembles of relatively shallow
  networks.
\newblock In D.~Lee, M.~Sugiyama, U.~Luxburg, I.~Guyon, and R.~Garnett,
  editors, \emph{Advances in Neural Information Processing Systems}, volume~29,
  pages 550--558. Curran Associates, Inc., 2016.
\newblock URL
  \url{https://proceedings.neurips.cc/paper/2016/file/37bc2f75bf1bcfe8450a1a41c200364c-Paper.pdf}.

\bibitem[Srivastava et~al.(2015)Srivastava, Greff, and
  Schmidhuber]{DBLP:journals/corr/SrivastavaGS15}
Rupesh~Kumar Srivastava, Klaus Greff, and J{\"{u}}rgen Schmidhuber.
\newblock Highway networks.
\newblock \emph{CoRR}, abs/1505.00387, 2015.
\newblock URL \url{http://arxiv.org/abs/1505.00387}.

\bibitem[He et~al.(2016{\natexlab{a}})He, Zhang, Ren, and
  Sun]{DBLP:conf/cvpr/HeZRS16}
Kaiming He, Xiangyu Zhang, Shaoqing Ren, and Jian Sun.
\newblock Deep residual learning for image recognition.
\newblock In \emph{2016 {IEEE} Conference on Computer Vision and Pattern
  Recognition, {CVPR} 2016, Las Vegas, NV, USA, June 27-30, 2016}, pages
  770--778. {IEEE} Computer Society, 2016{\natexlab{a}}.
\newblock \doi{10.1109/CVPR.2016.90}.
\newblock URL \url{https://doi.org/10.1109/CVPR.2016.90}.

\bibitem[He et~al.(2016{\natexlab{b}})He, Zhang, Ren, and
  Sun]{DBLP:conf/eccv/HeZRS16}
Kaiming He, Xiangyu Zhang, Shaoqing Ren, and Jian Sun.
\newblock Identity mappings in deep residual networks.
\newblock In Bastian Leibe, Jiri Matas, Nicu Sebe, and Max Welling, editors,
  \emph{Computer Vision - {ECCV} 2016 - 14th European Conference, Amsterdam,
  The Netherlands, October 11-14, 2016, Proceedings, Part {IV}}, volume 9908 of
  \emph{Lecture Notes in Computer Science}, pages 630--645. Springer,
  2016{\natexlab{b}}.
\newblock \doi{10.1007/978-3-319-46493-0\_38}.
\newblock URL \url{https://doi.org/10.1007/978-3-319-46493-0\_38}.

\bibitem[Larsson et~al.(2017)Larsson, Maire, and
  Shakhnarovich]{DBLP:conf/iclr/LarssonMS17}
Gustav Larsson, Michael Maire, and Gregory Shakhnarovich.
\newblock Fractalnet: Ultra-deep neural networks without residuals.
\newblock In \emph{5th International Conference on Learning Representations,
  {ICLR} 2017, Toulon, France, April 24-26, 2017, Conference Track
  Proceedings}. OpenReview.net, 2017.
\newblock URL \url{https://openreview.net/forum?id=S1VaB4cex}.

\bibitem[Liao and Poggio(2016)]{DBLP:journals/corr/LiaoP16}
Qianli Liao and Tomaso~A. Poggio.
\newblock Bridging the gaps between residual learning, recurrent neural
  networks and visual cortex.
\newblock \emph{CoRR}, abs/1604.03640, 2016.
\newblock URL \url{http://arxiv.org/abs/1604.03640}.

\bibitem[Chen et~al.(2018)Chen, Rubanova, Bettencourt, and
  Duvenaud]{DBLP:conf/nips/ChenRBD18}
Tian~Qi Chen, Yulia Rubanova, Jesse Bettencourt, and David Duvenaud.
\newblock Neural ordinary differential equations.
\newblock In Samy Bengio, Hanna~M. Wallach, Hugo Larochelle, Kristen Grauman,
  Nicol{\`{o}} Cesa{-}Bianchi, and Roman Garnett, editors, \emph{Advances in
  Neural Information Processing Systems 31: Annual Conference on Neural
  Information Processing Systems 2018, NeurIPS 2018, December 3-8, 2018,
  Montr{\'{e}}al, Canada}, pages 6572--6583, 2018.
\newblock URL
  \url{https://proceedings.neurips.cc/paper/2018/hash/69386f6bb1dfed68692a24c8686939b9-Abstract.html}.

\bibitem[Ioffe and Szegedy(2015)]{DBLP:conf/icml/IoffeS15}
Sergey Ioffe and Christian Szegedy.
\newblock Batch normalization: Accelerating deep network training by reducing
  internal covariate shift.
\newblock In Francis~R. Bach and David~M. Blei, editors, \emph{Proceedings of
  the 32nd International Conference on Machine Learning, {ICML} 2015, Lille,
  France, 6-11 July 2015}, volume~37 of \emph{{JMLR} Workshop and Conference
  Proceedings}, pages 448--456. JMLR.org, 2015.
\newblock URL \url{http://jmlr.org/proceedings/papers/v37/ioffe15.html}.

\bibitem[{Paszke} et~al.(2019){Paszke}, {Gross}, {Massa}, {Lerer}, {Bradbury},
  {Chanan}, {Killeen}, {Lin}, {Gimelshein}, {Antiga}, {Desmaison}, {K{\"o}pf},
  {Yang}, {DeVito}, {Raison}, {Tejani}, {Chilamkurthy}, {Steiner}, {Fang},
  {Bai}, and {Chintala}]{2019arXiv191201703P}
Adam {Paszke}, Sam {Gross}, Francisco {Massa}, Adam {Lerer}, James {Bradbury},
  Gregory {Chanan}, Trevor {Killeen}, Zeming {Lin}, Natalia {Gimelshein}, Luca
  {Antiga}, Alban {Desmaison}, Andreas {K{\"o}pf}, Edward {Yang}, Zach
  {DeVito}, Martin {Raison}, Alykhan {Tejani}, Sasank {Chilamkurthy}, Benoit
  {Steiner}, Lu~{Fang}, Junjie {Bai}, and Soumith {Chintala}.
\newblock {PyTorch: An Imperative Style, High-Performance Deep Learning
  Library}.
\newblock \emph{arXiv e-prints}, art. arXiv:1912.01703, December 2019.

\bibitem[Schoppe et~al.(2016)Schoppe, Harper, Willmore, King, and
  Schnupp]{10.3389/fncom.2016.00010}
Oliver Schoppe, Nicol~S. Harper, Ben D.~B. Willmore, Andrew~J. King, and Jan
  W.~H. Schnupp.
\newblock Measuring the performance of neural models.
\newblock \emph{Frontiers in Computational Neuroscience}, 10:\penalty0 10,
  2016.
\newblock ISSN 1662-5188.
\newblock \doi{10.3389/fncom.2016.00010}.
\newblock URL
  \url{https://www.frontiersin.org/article/10.3389/fncom.2016.00010}.

\bibitem[Hsu et~al.(2004)Hsu, Borst, and Theunissen]{Hsu:2004ku}
Anne Hsu, Alexander Borst, and Fr{\'e}d{\'e}ric Theunissen.
\newblock {Quantifying variability in neural responses and its application for
  the validation of model predictions}.
\newblock \emph{Network: Computation in Neural Systems}, 15\penalty0
  (2):\penalty0 91--109, May 2004.
\newblock \doi{10.1088/0954-898X/15/2/002}.
\newblock URL
  \url{http://www.informaworld.com/openurl?genre=article&doi=10.1088/0954-898X/15/2/002&magic=crossref||D404A21C5BB053405B1A640AFFD44AE3}.

\bibitem[Zipser et~al.(1996)Zipser, Lamme, and Schiller]{Zipser7376}
Karl Zipser, Victor A.~F. Lamme, and Peter~H. Schiller.
\newblock Contextual modulation in primary visual cortex.
\newblock \emph{Journal of Neuroscience}, 16\penalty0 (22):\penalty0
  7376--7389, 1996.
\newblock ISSN 0270-6474.
\newblock \doi{10.1523/JNEUROSCI.16-22-07376.1996}.
\newblock URL \url{https://www.jneurosci.org/content/16/22/7376}.

\bibitem[Hubel and Wiesel(1965)]{doi:10.1152/jn.1965.28.2.229}
David~H. Hubel and Torsten~N. Wiesel.
\newblock Receptive fields and functional architecture in two nonstriate visual
  areas (18 and 19) of the cat.
\newblock \emph{Journal of Neurophysiology}, 28\penalty0 (2):\penalty0
  229--289, 1965.
\newblock \doi{10.1152/jn.1965.28.2.229}.
\newblock URL \url{https://doi.org/10.1152/jn.1965.28.2.229}.
\newblock PMID: 14283058.

\bibitem[Allman et~al.(1985)Allman, Miezin, and
  McGuinness]{doi:10.1146/annurev.ne.08.030185.002203}
John Allman, Francis Miezin, and EveLynn McGuinness.
\newblock Stimulus specific responses from beyond the classical receptive
  field: Neurophysiological mechanisms for local-global comparisons in visual
  neurons.
\newblock \emph{Annual Review of Neuroscience}, 8\penalty0 (1):\penalty0
  407--430, 1985.
\newblock \doi{10.1146/annurev.ne.08.030185.002203}.
\newblock URL \url{https://doi.org/10.1146/annurev.ne.08.030185.002203}.
\newblock PMID: 3885829.

\bibitem[Cavanaugh et~al.(2002)Cavanaugh, Bair, and Movshon]{cavanaugh2002}
James~R. Cavanaugh, Wyeth Bair, and J.~Anthony Movshon.
\newblock Nature and interaction of signals from the receptive field center and
  surround in macaque v1 neurons.
\newblock \emph{Journal of Neurophysiology}, 88\penalty0 (5):\penalty0
  2530--2546, 2002.
\newblock \doi{10.1152/jn.00692.2001}.
\newblock URL \url{https://doi.org/10.1152/jn.00692.2001}.
\newblock PMID: 12424292.

\bibitem[Lurz et~al.(2021)Lurz, Bashiri, Willeke, Jagadish, Wang, Walker,
  Cadena, Muhammad, Cobos, Tolias, Ecker, and Sinz]{lurz2021generalization}
Konstantin-Klemens Lurz, Mohammad Bashiri, Konstantin Willeke, Akshay Jagadish,
  Eric Wang, Edgar~Y. Walker, Santiago~A Cadena, Taliah Muhammad, Erick Cobos,
  Andreas~S. Tolias, Alexander~S Ecker, and Fabian~H. Sinz.
\newblock Generalization in data-driven models of primary visual cortex.
\newblock In \emph{International Conference on Learning Representations}, 2021.
\newblock URL \url{https://openreview.net/forum?id=Tp7kI90Htd}.

\end{thebibliography}
